\documentclass[final,twocolumn,5p,times,10pt]{elsarticle}

\usepackage[usenames]{color}
\usepackage{amsfonts}
\usepackage{amssymb}
\usepackage{amsbsy}
\usepackage{multirow}
\usepackage{color}
\usepackage{times}
\usepackage{lscape}
\usepackage{graphicx}
\usepackage[cmex10]{amsmath}
\usepackage[english]{babel}
\usepackage[T1]{fontenc}
\usepackage[utf8]{inputenc}

\newtheorem{definition}{Definition}
\newtheorem{theorem}{\bf Theorem}

\newtheorem{property}{\it Property}

\newcommand{\w}{{\boldsymbol w}}
\newcommand{\Real}{\mathbb R}
\newcommand{\IG}{\includegraphics}

\newcommand{\fns}{\footnotesize}

\journal{Digital Signal Processing}

\begin{document}

\begin{frontmatter}

\title{{A Unified SVM Framework for Signal Estimation}}

\author[URJC]{Jos\'{e}~Luis Rojo-\'{A}lvarez}
\author[UNM]{Manel~Mart\'{i}nez-Ram\'on}
\author[UV]{Jordi~Mu\~noz-Mar\'{i}}
\author[UV]{Gustavo~Camps-Valls}

\address[URJC]{Dept. Teor\'{i}a de la Se\~nal y Comunicaciones,
  Universidad Rey Juan Carlos\\
  28943 Fuenlabrada, Madrid, Spain.
  {\it joseluis.rojo@urjc.es} }
  
\address[UNM]{Department of Electrical and Computer Engineering,
  University of New Mexico\\
  Albuquerque, NM 87131-0001 USA.
  {\it manel@ece.unm.edu}}
  
\address[UV]{Image Processing Laboratory (IPL),
  Universitat de Val\`encia \\
  46100 Burjassot, Val\`encia, Spain.
  {\it \{jordi.munoz,gustavo.camps\}@uv.es }}

\begin{abstract}
{This paper presents a review in the form of a unified framework for tackling estimation problems in Digital Signal Processing (DSP) using Support Vector Machines (SVMs). The paper formalizes our developments in the area of DSP with SVM principles. The use of SVMs for DSP  is already mature, and has gained popularity in recent years due to its advantages over other methods: SVMs are flexible non-linear methods that are intrinsically regularized and work well in low-sample-sized and high-dimensional problems. SVMs can be designed to take into account different noise sources in the formulation and to fuse heterogeneous information sources. Nevertheless, the use of SVMs in estimation problems has been traditionally limited to its mere use as a black-box model. Noting such limitations in the literature, } we take advantage of several properties of Mercer's kernels and functional analysis to {develop a family of SVM methods for estimation in DSP. Three types of signal model equations are analyzed. First, when a specific time-signal structure is assumed to model the underlying system that generated the data,} the linear signal model (so called {\it Primal Signal Model formulation}) is first stated and analyzed. Then, non-linear versions of the signal structure can be readily developed by following two different approaches. On the one hand, the signal model equation is written in reproducing kernel Hilbert spaces (RKHS) using the well-known {\it RKHS Signal Model formulation}, and Mercer's kernels are readily used in SVM non-linear algorithms. On the other hand, in the alternative and not so common {\it Dual Signal Model formulation}, a signal expansion is made by using an auxiliary signal model equation given by a non-linear regression of each time instant in the observed time series. {These building blocks can be used to generate different novel SVM-based methods for problems of signal estimation, and we deal with several of the most important ones in DSP.} We illustrate the usefulness of this methodology by defining SVM algorithms for linear and non-linear system identification, spectral analysis, nonuniform interpolation, sparse deconvolution, and array processing. The performance of the developed SVM methods is compared to standard approaches in all these settings. The experimental results illustrate the generality, simplicity, and capabilities of the proposed SVM framework for DSP.
\end{abstract}


\end{frontmatter}

\tableofcontents

\section{Introduction}	\label{introduction}

Digital Signal Processing (DSP) is a consolidated and active research area mainly devoted to detection, estimation, and time series analysis~\cite{Scharf90}. Among the numerous DSP applications, {\em detection algorithms} are widely applied to fields like sonar and radar detection, communication receivers, or speech recognition, whereas {\em estimation algorithms} are widely used for linear and non-linear plant or communication channel identification, estimation of angle of arrival in antenna arrays, and parametrization in speech coding and recognition. In addition, {\em time series algorithms} are widely used for stochastic systems control, forecasting, and spectrum analysis.

{Standard models in DSP have traditionally relied on the rather simplifying and strong assumptions of linearity, Gaussianity, stationarity, circularity, causality and uniform sampling. These models provide mathematical tractability and simple and fast algorithms, but they also limit the performance and applicability of these models. Since the 1980s, however, DSP has faced a dramatic change in model design. Current approaches try to get rid of these approximations, widely used models are intrinsically non-linear and nonparametric, and they can encode the relations between the signal and noise (which is often modeled and no longer considered Gaussian i.i.d. noise).
These issues have been fundamentally treated with non-linear models, such as neural networks. In the last decade, the field of DSP has witnessed the irruption and wide adoption of kernel methods in general and support vector machines (SVMs) in particular for all the aforementioned tasks.}

{SVM were originally conceived} as efficient methods for pattern recognition and classification~\cite{Vapnik95}, and the Support Vector Regressor (SVR) was subsequently proposed as the SVM implementation for regression and function approximation~\cite{Smola04,ShaweTaylor04}. Right after their introduction, researchers have applied it to a number of linear~\cite{Rojo05a} and non-linear DSP applications, such as speech recognition~\cite{Picone06}, image processing~\cite{Cremers03, Kim05}, channel equalization~\cite{Chen00}, multiuser detection~\cite{Chen01a,Chen01b,Bai03,Mati04}, array processing~\cite{Santamaria04a, Santamaria07}, or microwave design~\cite{Ayestaran05}. Adaptive SVM detectors and estimators for communication system applications have been also introduced~\cite{Navia01}. Beyond the SVM formulations, many other algorithms for DSP have also been stated from Mercer's kernel principles, with representative examples such as discriminant analysis~\cite{Mika99a,Baudat00}, clustering~\cite{BenHur01}, principal or independent component analysis~\cite{Hyvarinen01,Jordan02}, or mutual information extraction~\cite{Gretton03}.

SVMs have become a mature and recognized tool in DSP, the widespread adoption of SVM by researchers and practitioners in DSP being a direct consequence of their {good performance in terms of accuracy, sparsity, and flexibility. Note that SVMs are intrinsically regularized models implementing the maximum margin concept, they provide a natural way to perform data selection by choosing the most relevant vectors from a dataset (the so-called) support vectors, and can be engineered to accommodate different sources of information in the model.} The analysis of time series with supervised SVM algorithms has paid attention mainly to two DSP applications, namely, non-linear system identification and time series prediction~\cite{Drezet98,Gretton01b,Suykens01c,Suykens01d,Mattera04,Goethals05}. {In both problems, however, the SVM algorithm} was the conventional SVR using lagged samples of the available time signals as input vectors. Although good results have been reported with this approach, several concerns can be raised from a conceptual viewpoint of Estimation Theory:
\begin{enumerate}
\item The basic assumption for the regression problem statement, {in a Least Squares (LS) sense,} is that observations are independent and identically distributed ({\it i.i.d.}). This assumption of independence between samples is not fulfilled in time series data. Algorithms that do not take into account temporal dependences can miss relevant structures of the analyzed time signals, such as their autocorrelation or their cross-correlation. 
\item Most of these approaches use Vapnik's $\varepsilon$-insensitive cost function, which {linearly penalizes errors larger than $\varepsilon$ only. This is not the most appropriate loss function in the case of Gaussian noise in the data, which is the most common case in DSP problems.} 
\item These methods take advantage of the \emph{kernel trick}~\cite{Aizerman64} to develop non-linear versions from well established linear DSP techniques. However, the SVM methodology has other advantages which are desirable in DSP. {For instance, SVMs are intrinsically regularized algorithms that, unlike LS methods, are quite resistant to overfitting and robust in environments with low number of available training samples and high dimensional datasets. SVMs produces also sparse solutions provided by the used cost function which is advantageous for model interpretability and computational efficiency. SVMs also involve few model parameters to be tuned and lead to convex optimization problems unlike other popular models in DSP as neural networks. SVMs algorithms are founded on a solid mathematical background, hence bounds of performance and optimality conditions can be established. Actually, SVMs can benefit from the theory of reproducing kernel functions to, as we will see in this paper, treat heterogeneous information in a unified way.} 
\end{enumerate}

\begin{table*}[t]
\caption{{Scheme of the DSP-SVM framework (I): Equations of the time-series models for signal estimation.}}
{\footnotesize
\begin{center}
\begin{tabular}{|c|c|c|c|c|c|} \hline
\multirow{2}{*}{ } &  \multirow{2}{*}{Regression} &  \multicolumn{2}{c|}{Time-global } & \multicolumn{2}{c|}{Time-local}	 \\ \cline{3-6}
  	&		& Spectral  & ARx	& Sinc interp. & Deconv. \\		\hline
PSM & $\hat y = \langle \boldsymbol w, \boldsymbol x\rangle + b$ 	
	& $\hat y_n = \sum_{k=0}^K a_k cos(k\omega_0 t_n+\phi_k)$
	& $\hat y_n = \sum_{p=1}^Q D_p y_{n-p}+\sum_{q=0}^QE_qx_{n-q+1}$	
	& $\hat y_n = \sum_{k=0}^N a_k sinc(t-t_k)$	
	& $\hat y_n = x_n * h_n$  \\		\hline
RSM & $\hat y = \langle \boldsymbol w, \boldsymbol \varphi (\boldsymbol x)\rangle + b$	
	& -- &$\hat y_n = \langle \boldsymbol w_d, \boldsymbol \varphi_d(\boldsymbol y_n) \rangle + 
	\langle \boldsymbol w_e, \boldsymbol \varphi_e(\boldsymbol x_n) \rangle
	$ & --	& --	\\ \hline
DSM &  $\hat y = \sum_{i=1}^n \eta_i K(\boldsymbol x_i, \boldsymbol x) + b$
	 	& -- & -- & $ \hat y_n =  K(t) * \sum_k\eta_k \delta(t-t_k)$	
	  & $\hat y_n = \eta_n * R^h_n$   \\		\hline
\end{tabular}
\end{center}}
\label{tab:scheme1}
\end{table*}%

In recent years, several SVM algorithms for DSP applications have been proposed aiming to overcome the aforementioned limitations. A first approach to nonparametric spectral analysis, using the robust SVM optimization criterion instead of LS, was introduced in~\cite{Rojo03c}, where the robustness of the SVM against non-Gaussian noise was specifically addressed and solved. Afterwards, the robustness properties of the SVM were further exploited by proposing linear approaches for $\gamma$ filtering, ARMA modeling, array beamforming~\cite{Camps04,Rojo04,Martinez05}, and subspace-based spectrum estimation~\cite{Elgonnouni2012}. The non-linear generalization of ARMA filters with kernels~\cite{Martinez06a}, and temporal and spatial reference antenna beamforming using kernels and SVM~\cite{Martinez07}, have also been proposed. The use of convolutional signal mixtures has been addressed for interpolation and sparse deconvolution problems~\cite{Rojo07a,Rojo08a}, thanks to the autocorrelation kernel concept, a straightforward property which has opened the field for a number of unidimensional and multidimensional extensions of communications problems~\cite{Figuera2012,Figuera2013}. {These are examples that represent partial contributions to the more general problem of building non-linear SVMs to tackle DSP problems.} Other recent works make use of {reproducing kernel Hilbert spaces (RKHS)} signal model equations, but not using SVM optimization (see, e.g.~\cite{Arenas2013, Perez-Cruz2013, Ding2013}).

\section{ {Rationale and Structure of the Review}} \label{sec:rationale}

\begin{table}[t]
\caption{{Scheme of the DSP-SVM framework (II): Equations for signal models in array processing.}}
{\footnotesize
\begin{center}
\begin{tabular}{|c|c|c|} \hline
 	 		&  \multicolumn{2}{c|}{Antenna Array Processing} \\ \cline{2-3}
  			   & Temporal reference  			& Spatial reference \\	\hline
PSM  &  $\hat y_n = \langle \boldsymbol a,\boldsymbol x_n\rangle$
	 & \multirow{2}{*}{  $ b_i = \langle \boldsymbol  w, \boldsymbol \varphi (b_i \boldsymbol a_0)\rangle - b$ } \\ \cline{1-2} 
RSM & $\hat y_n = \langle \boldsymbol w, \boldsymbol \varphi (\boldsymbol x_n)\rangle$	&  \\ \hline
DSM & -- & --  \\	\hline
\end{tabular}
\end{center}}
\label{tab:scheme2}
\end{table}%

This paper provides a landscape of the preceding works, and the formalization of a unified framework for developing SVM algorithms for {supervised estimation applications in DSP}. The framework is thus focused on time series analysis, in which the time structure of the data could be highly informative. We start from the consideration that discrete-time processes should be treated in a conceptually different way from a regression model, to efficiently deal with data with underlying time series structure. {This framework can be summarized as follows:} 
\begin{itemize}
\item The statement of linear signal model equations in the primal problem, or {\it SVM Primal Signal Models} (PSM), allows us to obtain robust estimators of the model coefficients~\cite{Rojo05a} and to take advantage of almost all the characteristics of the SVM methodology in classical DSP problems, such as ARX time series modeling, spectral analysis~\cite{Rojo03c, Rojo04,Camps04}, and {antenna array signal processing}~\cite{Martinez07}.
\item The first option for the statement of non-linear signal model equations are the widely used {\it RKHS Signal Models} (RSM), which state the signal model equation in the RKHS and substitute the dot products by Mercer's kernels~\cite{Vapnik95,Martinez06a,Martinez07}.
\item The second option are the {\it Dual Signal Models} (DSM), which have been previously proposed in an implicit way, and are based on the non-linear regression of the time instants with appropriate Mercer's kernels~\cite{Rojo07a,Rojo08a}. While RSM allow us to scrutinize the statistical properties in the RKHS, DSM can give an interesting and straightforward interpretation of the SVM algorithm under study, in connection with classical Linear System Theory.
\end{itemize}
{This framework is summarized in Table \ref{tab:scheme1}, where the key regression equations are shown for better understanding and handling of the signal model equations. The SVM regression formulation is well known and widespread, and it is included for the sake of completeness and tutorial purposes.
The table gives a principled framework for building efficient SVM linear and non-linear algorithms in DSP applications. The provided algorithms makes use of these three types of signal model equations, which can consider the time series structure of the data in different ways.} 

{Note that there is an almost endless variety of signal model equations in DSP. Among them, we choose the following ones:
\begin{itemize}
\item From a viewpoint of the nature of the signals that can be used, we consider time-global and time-local signal expansions. The former are given by basis signals whose duration expands in a non-decaying way throughout the time interval where the estimated signal is observed. In particular, sinusoids in nonparametric spectral estimation, and delayed versions of exogenous and endogenous signals in difference equation models. The later are given by basis signals which are either duration-limited or decaying, which is the case of sinc functions in time series interpolation, or energy-defined impulse responses in deconvolution problems. Illustrative examples of these kinds of equations are summarized in Table \ref{tab:scheme1}.
\item A different, yet related approach comes when different applications can be stated according to different unknown terms of the same specific signal model equation. An excellent example in this setting is antenna array processing for beamforming, where the same signal model equation supports temporal reference signal detection, and spatial reference estimation problems. As illustrated in Table \ref{tab:scheme2}, PSM and RSM have been proposed for temporal reference problems, in a similar way than DSP problems in Table \ref{tab:scheme1}, but interestingly, a slightly different signal model is used in spatial reference, expressed in both cases in terms of possibly nonlinear mapping. This aims to illustrate that, in this case, we switch to a eigenproblem statement, which is a better representation for the data model, both for linear and nonlinear cases.
\end{itemize}
Equations in Tables \ref{tab:scheme1} and \ref{tab:scheme2} will be explained in detail throughout the paper, so the reader is encouraged to come back to these tables after the first reading. Note also that, in these tables, many problems have not been addressed yet (as indicated by "--"), and our intention is to motivate the interested reader to complete and expand this table according to their own DSP needs.
}

The remainder of the paper is as follows. In the next section, the well-known elements of the non-linear SVR algorithm are briefly summarized, as they contain all the fundamental tools that will be required for estimation problems. In Section 3, a general signal model equation is proposed for supervised learning in time series estimation, and several signal model equations from representative DSP fields are introduced accordingly.  In Section 4, the PSM Theorem allows us to create linear algorithms for the described signal models. In Section 5, the RSM Theorem is explicitly stated, yielding non-linear algorithms for system identification and for sinusoid detection in the RKHS. In Section 6, the DSM Theorem allows an immediate formulation for convolutional data models, such as {\em sinc} nonuniform interpolation and sparse deconvolution. Finally, in Section 7, discussion and conclusions are given.

\section{SVM Elements for Regression and Estimation}	\label{elements}

The SVR algorithm contains all the key elements to tackle estimation problems in our setting, i.e. regularized solution, convexity of the optimization problem, {sparsity of the solution, flexibility of the non-linear model via kernel functions}, and adaptiveness to different noise sources.  Only the properties that are relevant to this paper are summarized here, but the interested reader can see detailed derivations available in the SVM literature (see e.g. the classical book by Cristianini~\cite{Cristianini99} and references therein).

\begin{definition}[Nonlinear SVR Signal Model Hypothesis] \label{prop:SVRmodel}
Be a labeled training {\it iid} data set ${\{({\boldsymbol v}_i,l_i)}$, ${i=1,...,N_r\}}$, where ${\boldsymbol v}_i\in {\mathbb{R}}^d$ and $l_i\in {\mathbb{R}}$. The SVR signal model first maps the observed explanatory vectors to a higher dimensional kernel feature space through a non-linear mapping $\boldsymbol{\phi}:{\mathbb R}^{N_r}\longrightarrow{\mathcal H}$,
and then obtains a linear regression model inside this space, this is,
\begin{equation}
{\hat l}_i = \langle \w, \boldsymbol{\phi}(\boldsymbol{v}_i) \rangle + b,
\end{equation}
where $\w$ is a weight vector in ${\mathcal H}$ and $b$ is the bias term in the regression. Model residuals are given by $e_i = l_i - {\hat l}_i $.
\end{definition}
In order to obtain the model coefficients, the SVR minimizes a cost function of the residuals, which is usually regularized by the $L_2$ norm of $\w$. This is, we minimize
\begin{equation}\label{eq:primalRaw}
 \frac{1}{2}\|\w\|^2 + \sum_{i=1}^{N_r} {\mathcal L}(e_i)
\end{equation}
In the standard SVR formulation, Vapnik's $\varepsilon$-insensitive cost is often used~\cite{Vapnik98}.

\begin{figure}[t] \centering
\begin{tabular}{cc}
\multicolumn{2}{c}{\IG[width=8cm]{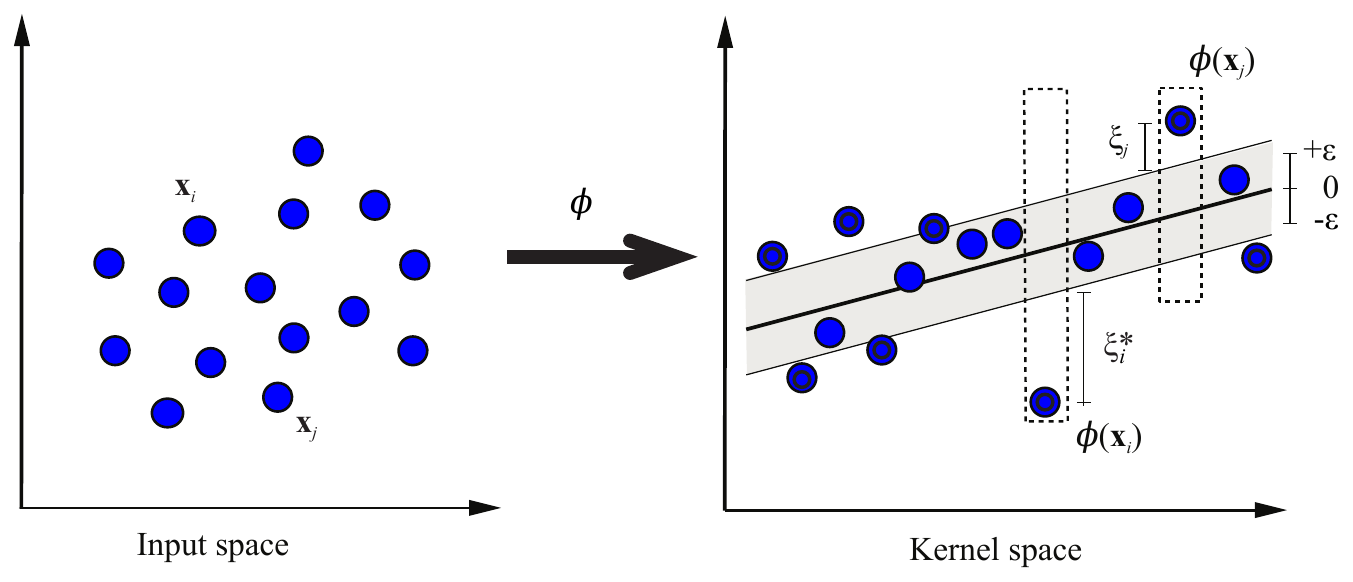}} \\
\multicolumn{2}{c}{\IG[width=8cm]{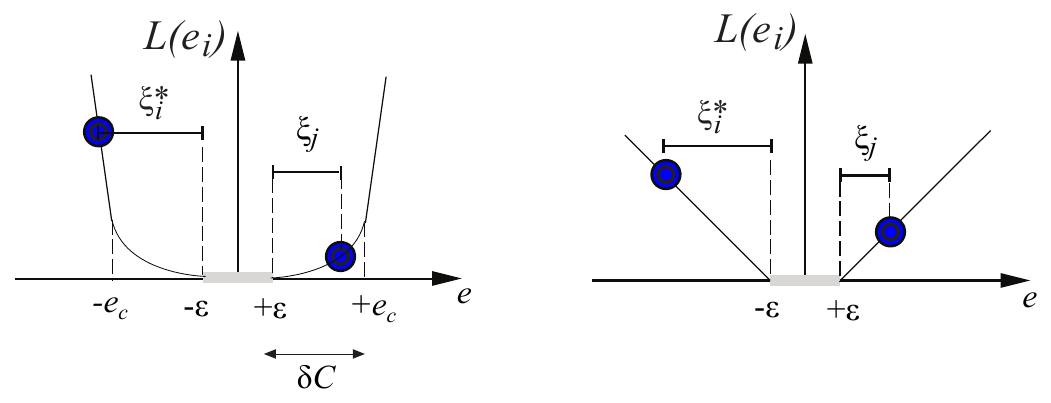}} \\
{\fns\it(a)} & {\fns\it(b)} \\
\end{tabular}
 \caption{SVR signal model. Samples in the original input space are first mapped to an RKHS where a linear regression is performed. All samples outside a fixed tube of size $\varepsilon$ are penalized, and are support vectors (double-circled symbols). Penalization is carried out by applying (a) Vapnik's $\varepsilon$-insensitive or (b) $\varepsilon$-Huber cost
 functions.}
 \label{svr}
\end{figure}

\begin{definition}[Vapnik's $\varepsilon$-insensitive Cost]
Given a set or residual errors $e_i$ in an estimation problem, the $\varepsilon$-insensitive cost is given by
\begin{eqnarray}
{\mathcal L}_{\varepsilon}(e_i) = C  \max(|e_i|-\varepsilon,0),
\end{eqnarray}
where $C$ controls the trade-off between the regularization and the losses. Residuals lower than $\varepsilon$ are not penalized, whereas larger ones have linear cost.
\end{definition}

This cost function is a suboptimal estimator in applications when combined with a regularization term~\cite{Vapnik95,Scholkopf02,Smola04} and the noise follows a Gaussian distribution, because of the linear nature of the cost function. 
This issue has been addressed in the formulation of LS-SVM~\cite{Suykens01}, where a quadratic cost is used, though in this case, sparseness of the solution is lost. An alternative cost function of the residuals, the $\varepsilon$-Huber cost, has been proposed~\cite{Rojo02b} by combining both the quadratic and the $\varepsilon$-insensitive cost. This has been shown to be a more appropriate residual cost for SVR in general~\cite{Camps06a}.
\begin{definition}[$\varepsilon$-Huber Cost Function]
The $\varepsilon$-Huber cost is given by
\begin{equation}\label{eq:ehuber}
{\mathcal L}_{\varepsilon H}(e_i)=
 \begin{cases}
 0, & |e_i|\leq\varepsilon \\
 \frac{1}{2 \delta}(|e_i| -\varepsilon)^2, & \varepsilon \leq |e_i| \leq e_C \\
 C(|e_i|-\varepsilon)- \frac{1}{2} \delta C^2, & |e_i| \geq e_C
 \end{cases}
\end{equation}
where $e_C=\varepsilon+\delta C$; $\varepsilon$ is the insensitive parameter, and $\delta$ and $C$
control the trade-off between the regularization and the losses.
\end{definition}
The three different regions in the $\varepsilon$-Huber cost allow us to deal with different kinds of noise: the $\varepsilon$-insensitive zone ignores absolute residuals lower than $\varepsilon$; the quadratic cost zone uses the $L_2$-norm of the residuals, which is appropriate for Gaussian noise; and the linear cost zone is an efficient limit for the impact of the outliers in the optimal model coefficients. Note that~\eqref{eq:ehuber} represents Vapnik's $\varepsilon$-insensitive cost function when $\delta$ is small enough, LS criterion for $\delta C \rightarrow \infty$ and $\varepsilon = 0$ , and Huber's cost function when $\varepsilon=0$ (see Fig.~\ref{svr}).

By including the $\varepsilon$-Huber cost into~\eqref{eq:primalRaw}, the estimation of the SVR coefficients can be obtained as the minimization of a Quadratic Programming (QP) problem~\cite{Smola04,Rojo04}. Several additional and very relevant properties can be obtained from the Karush-Khun-Tucker conditions and the dual functional, which are next summarized.

\begin{property}[SVR Sparse Solution and Support Vectors] 
The weight vector  in ${\mathcal H}$ can be expanded in a linear combination of the transformed input data,
\begin{equation}
\w = \sum_{i=1}^{N_r} \eta_i \boldsymbol{\phi}(\boldsymbol{v}_i), \label{eq:con_dual1}
\end{equation}
where $\eta_i = (\alpha_i - \alpha_i^\ast)$ are the model weights, and $\alpha_i^{(\ast)}$ are the Lagrange multipliers corresponding to the positive and negative residuals in the $i^{th}$ observation. {Observations with non-zero associated coefficients are called {\emph support vectors}, and the solution is expressed as a function of them solely.}
\end{property}

\begin{property}[Robust Expansion Coefficients]\label{property:expansioncoefficients}
The following non-linear relationship between the residuals {and the model coefficients for the $\varepsilon$-Huber cost is given by}:
\begin{equation}\label{eq:KKTresid}
\begin{split}
\eta_i &=  \frac{\partial L_{\varepsilon H} (e) }{\partial e}\bigg|_{{e = e_i}} \\
&=
\begin{cases}
 0, & |e_i|\leq \varepsilon \\
\frac{1}{\delta} \cdot sign ({e_i})( |e_i| - \varepsilon),& \varepsilon < |e_i| \leq \varepsilon + \gamma C\\
C \cdot sign ({e_i}) , & |e_i| > \varepsilon + \gamma C\\
\end{cases}
\end{split}
\end{equation}
Therefore, the impact of a large residual $e_i$ on the coefficients is limited by the value of $C$ in the cost function, which yields estimates of the model coefficients that are robust in the presence of outliers.
\end{property}

\begin{theorem}[Mercer's Theorem~\cite{Aizerman64}] Let $K(\boldsymbol{u}, \boldsymbol{v})$ be a bivariate function fulfilling the Mercer condition, i.e., $\int_{\mathbb{R}^{N_r}\times \mathbb{R}^{N_r}} K(\boldsymbol{u}, \boldsymbol{v}) f(\boldsymbol{u})f(\boldsymbol{v}) \geq 0$ for any square integrable function $f(\boldsymbol{u})$. 
Then, there exists a Reproducing Kernel Hilbert Space (RKHS) ${{\mathcal H}}$ and a mapping ${\boldsymbol \phi}(\cdot)$, such that $K(\boldsymbol{u}, \boldsymbol{v}) = \langle {\boldsymbol \phi} (\boldsymbol{u}),{\boldsymbol \phi}(\boldsymbol{v}) \rangle$ . 
\end{theorem}
The kernel trick in SVM consists of stating the problem at hand (e.g. classification, regression, and many others) in terms of dot products of data in the RKHS, and then substituting these products by Mercer's kernels. The kernel expression is actually used in a given SVM algorithm, but neither the mapping function ${\boldsymbol \phi}(\cdot)$, nor the RKHS, need to be known explicitly. 
{The Lagrangian of~\eqref{eq:primalRaw} is used to obtain the dual problem, which in turn yields the Lagrange multipliers used as model coefficients.}

The following Mercer's kernels are often used in SVM literature:
\begin{eqnarray}
K({\boldsymbol{u}},{\boldsymbol{v}}) & = & \langle{\boldsymbol{u}},{\boldsymbol{v}}\rangle \\
K({\boldsymbol{u}},{\boldsymbol{v}}) & = & (\langle {\boldsymbol{u}},{\boldsymbol{v}}\rangle+1)^d \\
K({\boldsymbol{u}},{\boldsymbol{v}}) & = & \exp \left(-\|{\boldsymbol{u}} - {\boldsymbol{v}}\|^2/2\sigma^2 \right)
\end{eqnarray}
which are called linear, polynomial ($d\in {\mathbb Z}^+$), and Radial Basis Function (RBF) ($\sigma\in \Real^+$) Mercer's kernels, respectively.

\begin{property}[Regularization in the Dual]
The dual problem of~\eqref{eq:primalRaw} for the $\varepsilon$-Huber cost corresponds to the maximization of
 \begin{equation}
 -\frac{1}{2} (\boldsymbol{\alpha} - \boldsymbol{\alpha}^\ast)^\top \left({\boldsymbol K} + \delta {\boldsymbol I}\right)
 (\boldsymbol{\alpha} - \boldsymbol{\alpha}^\ast) + (\boldsymbol{\alpha} - \boldsymbol{\alpha}^\ast)^\top \boldsymbol{l}
 -\varepsilon \boldsymbol{1}^\top (\boldsymbol{\alpha} + \boldsymbol{\alpha}^\ast)
\end{equation}
constrained to $0 \leq \alpha_i^{(\ast)} \leq C$. Here, $\boldsymbol{\alpha}^{(\ast)}$ = $[ \alpha_1^{(\ast)},
\cdots, \alpha_{n}^{(\ast)}]^\top$; $\boldsymbol{l} = [l_{1},\cdots,l_{N_r}]^\top$; ${\boldsymbol K}$ represents
the kernel matrix, given by ${\boldsymbol K}_{i,j} = K({\boldsymbol v}_i,{\boldsymbol v}_j) = \langle \boldsymbol{\phi}({\boldsymbol v}_i),\boldsymbol{\phi}({\boldsymbol v}_j) \rangle$; $\boldsymbol{1}$ is an all-ones column vector; and ${\boldsymbol I}$ is the identity matrix. 
\end{property}
The use of the quadratic zone in the $\varepsilon$-Huber cost function gives rise to a numerical regularization. The effect of $\delta$ in the solution is analyzed in~\cite{Rojo04}.
\begin{property}[Estimator as an Expansion of Kernels]\label{property:ExpansionKernels}
The estimator is given by a linear regression in the RKHS, and it can be expressed only in terms of the Lagrange multipliers and Mercer's kernels as 
\begin{equation}
{\hat l} ({\boldsymbol v}) = \langle {\boldsymbol w},{\boldsymbol \phi}({\boldsymbol v})\rangle + b = \sum_{i=1}^{N_r} \eta_i K({\boldsymbol v}_i,{\boldsymbol v}) + b,
\end{equation}
where only the support vectors (i.e. training examples whose corresponding Lagrange multipliers are non-zero) contribute to the solution.
\end{property}

\section{{Signal Processing Problems and their Signal Models}}	\label{signalmodels}

{The proposed SVM framework for DSP} consists of several basic tools and procedures. We start by defining a general signal model equation for considering a time series structure in our observed data, consisting on an expansion in terms of a set of signals spanning a Hilbert signal subspace and a set of model coefficients to be estimated. {Then this general signal model equation is specified for the chosen DSP problems according to the rationale given in Section \ref{sec:rationale}.}

\begin{definition}[General Signal Model Hypothesis] \label{prop:general} 
Given a time series $\{y_n\}$ consisting of $N+1$ observations, with $n=0,\dots,N$, an expansion for approximating this signal can be built with a set of signals $\{s_n^{(k)}\}$, with $k=0,\dots,K$, spanning the Hilbert signal subspace. This expansion is given by
\begin{equation}\label{eq:expansion}
 {\hat y}_n = \sum_{k=0}^K a_k s_n^{(k)},
\end{equation}
where $a_k$ are the expansion coefficients, to be estimated according to some adequate criterion, and $e_n = y_n - {\hat y}_n$ are the model residuals.
\end{definition}
The set $\{s_n^{(k)}\}$ are the {\it explanatory signals}, {which are selected here to encode the {\it a priori} belief about the time-series structure} of the observations. After that, the estimation of the expansion coefficients has to be addressed.

\begin{definition}[Optimization Functional] Given a signal to be modeled and a set of explanatory signals, an optimization functional is used to estimate model coefficients. The functional is a linear combination of a loss  $\mathcal{L}$ of  residuals  $e_n$, and a regularization functional $\mathcal{M}$ (e.g., Tikhonov regularizer~\cite{Tikhonov77}) expressed in terms of  estimated coefficients $a_k$, this is,
\begin{equation}
 \{a_k^{opt}\} = \arg_{opt} \Big{\{} \sum_{n=0}^N \mathcal{L}(e_n) + \mathcal{M}(a_k) \Big{\}}.
\end{equation}
\end{definition}

Therefore, a general problem on time series modeling consists on first looking for an adequate set of explanatory signals, and then estimating the coefficients with a proper criterion for the residuals and for these coefficients. Several  signal model equations have been paid attention in the DSP and Information Theory literature,  whose signal structure is better analyzed by taking into account their correlation information. They have been previously addressed with the SVM methodology,  and we put them in a framework for indicating their differences and {common points} in the next subsections.

\subsection{Nonparametric Spectral Estimation}

In {\it Nonparametric Spectral Estimation,} the signal model hypothesis is a linear combination  of a set explanatory signals which are sinusoidal waveforms, from a given grid of frequencies and with amplitudes and phases to be estimated.
When the signal to be spectrally analyzed is uniformly sampled, {the LS criterion yields methods based on the Fourier transform, such as the Welch's periodogram and the Blackman-Tukey's correlogram~\cite{Marple87}.} 
When the signal is non-uniformly sampled, the in-phase and quadrature-phase components  of the basis are still orthogonal at the uneven sampling times, thus yielding the Lomb periodogram~\cite{Lomb76}. 

\begin{property}[Sinusoidal Signal Model Hypothesis] \label{prop:sinusoidalModel}
Given a set of observations $\{y_n\}$, which is known to present a spectral structure, its signal model hypothesis can be stated as:
\begin{equation}
\begin{split}
 {\hat y}_n = & \sum_{k=0}^K a_k s_n^{(k)} = \sum_{k=0}^K A_k \cos(k \omega_0 t_n + \phi_k ) =\\
  & = \sum_{k=0}^K \left( B_k \cos(k \omega_0 t_n ) + C_k \sin(k \omega_0 t_n ) \right),
\end{split}
\end{equation}
where angular frequencies are assumed to be previously known or fixed in a regular grid with spacing $\omega_0$; $A_k, \phi_k$ are the amplitudes and phases of the $k^{th}$ components, and $B_k = A_k \cos(\phi_k)$ and $C_k = A_k \sin(\phi_k)$ are the in-phase and in-quadrature model coefficients, respectively; and $\{t_n\}$ are the (possibly unevenly separated) sampling time instants.
\end{property}
The Sinusoidal Signal Model straightforwardly corresponds to the General Signal Model in Definition~\ref{prop:general} for $\{a_k \} \equiv \{ B_k\} \cup \{C_k \} $ and $\{s_n^{(k)}\} \equiv \{ \sin(k \omega_0 t_n )\} \cup \{ \cos(k \omega_0 t_n ) \}$. {Additionally, note that this signal model equation allows us to consider the spectral analysis of continuous-time unevenly sampled time series.}

\subsection{ARX System Identification}

In {\it Parametric System Identification and Time Series Prediction,} the signal model hypothesis is driven by a difference equation, and the explanatory signals are delayed versions of the same observed signal, and possibly (for system identification) by delayed versions of an exogenous signal.
A common problem in DSP is to model a functional relationship between two simultaneously recorded discrete-time processes~\cite{Ljung99}. When this relationship is linear and time-invariant, it can be addressed using an Auto-Regressive and Moving Average (ARMA) difference equation. {When a simultaneously observed (exogenous) signal $\{x_n\}$ is available, the ARX signal model equation is used for system identification.}

\begin{property}[ARX Signal Model Hypothesis] \label{prop:ARXmodel}
Given a set of observations $\{y_n\}$, and a simultaneously observed signal $\{x_n\}$, an ARX signal model hypothesis can be stated between them in terms of a parametric model described by an ARMA difference equation, given by delayed versions of both,
\begin{equation}\label{eq:spectralSignalModel}
 {\hat y}_n = \sum_{k=0}^K a_k s_n^{(k)} = \sum_{p=1}^P D_p y_{n-p} + \sum_{q=0}^Q E_q x_{n-q},
\end{equation}
where $\{x_n\}$ is the exogenous signal; $D_p$ and $E_q$ are the AR and the X model coefficients, respectively, and the system identification is an ARX signal model equation.
\end{property}
The ARX System Identification Signal Model is the General Signal Model in Definition~\ref{prop:general} for $\{a_k\} \equiv \{ D_p\} \cup \{E_q \} $ and $\{s_n^{(k)}\} \equiv \{ y_{n-p}\} \cup \{x_{n-q} \}$.

\subsection{Sinc Kernel Interpolation}

In {\it Sinc Interpolation}, a band-limited signal model is hypothesized, and the explanatory signals are delayed sincs. The {\em sinc} kernel provides the perfect reconstruction of an evenly sampled noise-free signal~\cite{Oppenheim89}. In the presence of noise, the sinc reconstruction of a possibly non-uniformly sampled time series is an ill-posed problem~\cite{Yen56,Munson98,Munson00}. 

\begin{property}[Sinc Kernel Signal Model Hypothesis] \label{prop:sincmodel}
Let $y(t)$ be a band limited, possibly Gaussian noise corrupted signal, and be $\{y_k = y(t_k), k=0,\dots,N\}$ a set of $N+1$ nonuniformly sampled observations. The sinc interpolation problem consists of finding an approximating function ${\hat y}(t)$ fitting the data,   ${\hat y}(t) = \sum_{k=0}^N a_k \textrm{sinc}(\sigma_0(t-t_k))$.
The previous continuous time model, after non-uniform sampling, is expressed as the following discrete time model:
\begin{equation} \label{eq:ymodeldiscrete}
 y_n = {\hat y}_n + e_n = \sum_{k=0}^N a_k \textrm{sinc}(\sigma_0(t_n-t_k)) + e_n,
\end{equation}
where $\textrm{sinc}(t) = \frac{\sin(t)}{t}$, and $\sigma_0 = \frac{\pi}{T_0}$ is the sinc units bandwidth.
\end{property}
Therefore, we are using an expansion of {\em sinc} kernels for interpolation of the observed signal. The {\em sinc} kernel interpolation signal model straightforwardly corresponds to the General Signal Model in Definition~\ref{prop:general} for explanatory signals $\{s_n^{(k)}\} \equiv \{ \textrm{sinc}(\sigma_0(t_n-t_k)) \}$. An optimal band-limited interpolation algorithm, in the LS sense, was first proposed in~\cite{Yen56}.

\subsection{Sparse Deconvolution}

In {\it Sparse Deconvolution,} the signal hypothesis is given by a convolutional signal mixture, and the explanatory signals are the delayed and scaled versions of the impulse response from a previously known Linear Time Invariant system. More specifically, the sparse deconvolution problem  consists on the estimation of an unknown sparse sequence which has been convolved with a (known) time series (impulse response of the system or source wavelet) and corrupted by noise, hence producing the observed noisy time series. The non-null samples of the sparse series contain relevant information about the underlying physical phenomenon in each application. 

\begin{property}[Sparse Deconvolution Signal Model Hypothesis] \label{prop:sparsedeconmodel}
Let $\{y_n\}$ be a discrete-time signal given by $N+1$ observed samples of a time series, which is the result of the convolution between an unknown sparse signal $\{x_n\}$, to be estimated, and a known time series $\{h_n\}$ (with $M+1$ duration). Then, the following convolutional signal model equation can be written,
\begin{equation} \label{eq:dconv}
 {\hat y}_n = {\hat x}_n \ast h_n = \sum_{j=0}^{M} {\hat x}_j h_{n-j} = \sum_{k=0}^K a^{(k)}s_n^{(k)},
\end{equation}
where $\ast$ denotes the discrete-time convolution operator, and $\hat x_n$ is the estimation of the unknown input signal. 
\end{property}
The Sparse Deconvolution Signal Model corresponds to the General Signal Model in Definition~\ref{prop:general} for model coefficients $\{a_k\} \equiv \{ x_k \} $ and explanatory signals $\{s_n^{(k)}\} \equiv \{ h_{n-k} \}$. The performance of sparse deconvolution algorithms can be degraded by several causes. First, they can result in ill-posed problems~\cite{Santamaria96}, and regularization methods are often required. Also, when the noise is non-Gaussian, either LS or maximum likelihood deconvolution can yield suboptimal solutions~\cite{kassam85}. Finally, if ${h_n}$ has non-minimum phase, some sparse deconvolution algorithms needing inverse filtering become unstable.

\subsection{Array Processing}

In  {\it Array Processing}, a complex-valued spatio-temporal signal model equation is used in order to manage the properties of an array of antennas in several signal processing applications.
The easiest system model in array processing~\cite{VanTrees02} consists of a linear array of $K+1$ elements equally spaced a distance $d$, whose output is a time series of vector samples ${\boldsymbol x}_n=\{x^0_n, \cdots, x^K_n\}^T$ or snapshots. Usually, signals $x^k_n$ are represented as lowpass signals. In order to keep their amplitude and phase information, signals need a complex-valued representation in terms of in-phase and quadrature-phase components. A given source with constant amplitude, whose direction of arrival (DOA) and wavelength are $\theta_l$ and $\lambda$, respectively, yields the following array output (so-called {\it steering vector}),
\begin{equation}\label{eq:steering_vector}
{\boldsymbol a}_l=\{1,e^{j2\pi\frac{d}{\lambda}sin(\theta_l)},\cdots,e^{j 2 K \pi\frac{d}{\lambda}sin(\theta_l)}\}^\top,
\end{equation}
where $j=\sqrt{-1}$. If $L$ transmitters are present, the snapshot can be represented as
$
{\boldsymbol x}_n={\boldsymbol A}{\boldsymbol b}_n + {\boldsymbol n}_n
$
where ${\bf A}$ is a matrix containing all steering vectors of the $L$ transmitters, ${\boldsymbol b}_n$ is a column vector containing (complex valued) symbols transmitted by all users and ${\boldsymbol n}_n$ is the thermal noise present at the output of each antenna. 

{\begin{property}[Array Processing Signal Model Hypothesis] \label{prop:arraymodel}
Let $\{y_n\}$ be a discrete time signal given by $N-1$ symbols transmitted by user $0$, and be $x^k_n, 0\leq k\leq K$, a set of discrete time processes representing time samples of the measured current at each of the array elements. The array processing problem consists of estimating $y_n$ as
{
\begin{equation}\label{eq:array_problem}
\hat y_n=\sum_{k=0}^K a_k x^k_n. 
\end{equation}}
\end{property}}
The problem is called {\it array processing with temporal reference} estimation problem when a set of transmitted signals is previously observed for training purposes. Whenever the DOA of the desired user is known, the problem is an {\it array processing with spatial reference} problem. This signal model equation agrees with Definition~\ref{prop:general} for $\{s_n^{k}\} = \{x^k_n\}$.

\section{Type I Algorithms: Primal Signal Models}		\label{primal}

A class of linear {SVM algorithms for DSP} come from the so called PSM~\cite{Rojo05a}. In this framework, rather than the prediction of the observed signal, the goal of the SVM is a set of model coefficients that contain the relevant information. The use of the $\varepsilon$-Huber cost allows to deal with Gaussian noise in all the {SVM algorithms for DSP}, while still providing with robust estimations of the model coefficients. {As  illustrated in Tables \ref{tab:scheme1} and \ref{tab:scheme2}, this approach consists of using the signal model equations in the preceding section in a SVR-like linear formulation.}

\subsection{Fundamentals of PSM}

 \begin{figure}[t]
 \centering
 \IG[width=\columnwidth]{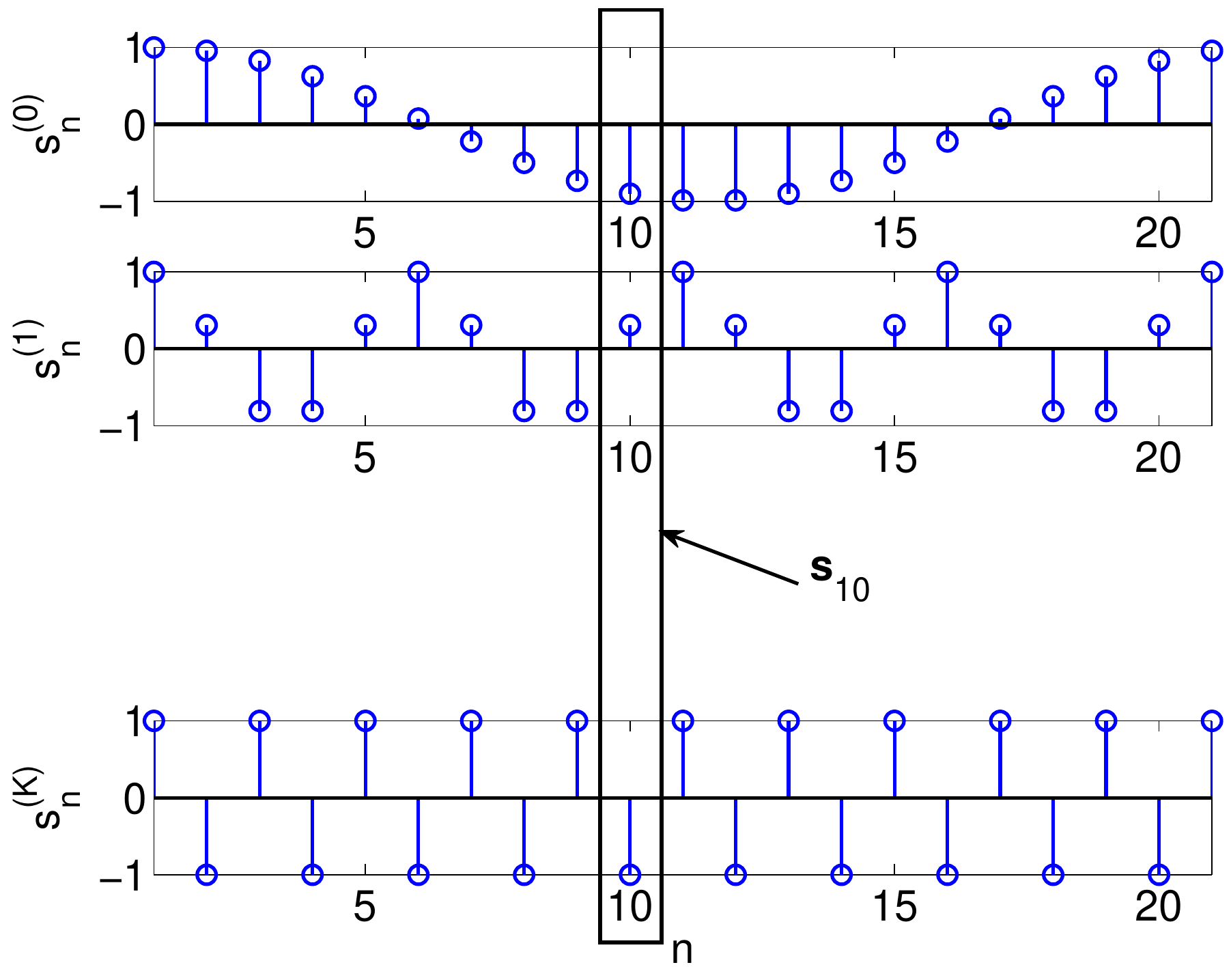}
 \caption{In this figure $K+1$, ($0 \leq k \leq K$) sinusoidal explanatory signals $s_n^{(k)}$ are represented. Time transversal vector ${\bf s}_{10}$ is constructed with the $10^{th}$ sample of each one of the explanatory signals. }
 \label{fig:transversal}
 \end{figure}

\begin{definition}[Time-Transversal Vector of a Signal Expansion]
Let $\{y_n\}$ be a discrete time series in a Hilbert space, and given the General Signal Model in Definition~\ref{prop:general}, then the $n^{th}$ time-transversal vector of the signals in the generating expansion set $\{ s_n^{(k)}\}$ is defined as
\begin{equation}
 \boldsymbol{s}_n = [ s_n^{(0)}, s_n^{(1)},\dots, s_n^{(K)}]^\top.
\end{equation}
Hence, it is given by the $n^{th}$ samples of each of the signals generating the signal subspace where the signal approximation is made. 
\end{definition}
Figure \ref{fig:transversal} depicts a pictorial example, in which time-transversal vector ${\boldsymbol s}_{10}$ is given by the $10^{th}$ sample of each explanatory signal (sinusoidal signals in the graph example).

Accordingly, the PSM problem can be stated as follows.

\begin{theorem}[PSM Problem Statement]
Let $\{y_n\}$ be a discrete time series in a Hilbert space,  then the optimization of
\begin{equation} 
 \frac{1}{2}\|{\boldsymbol a}\|^2 + \sum_{n=0}^N {\mathcal L}_{\varepsilon H}(e_n)
\end{equation}
with ${\boldsymbol a} = [ a_0, a_1,\dots, a_K]^\top$, gives an expansion solution whose signal model equation is
\begin{equation} \label{eq:primalsignalmodel}
 {\hat y}_n = \sum_{k=1}^K a_k s_n^{(k)} = \langle {\boldsymbol a}, {\boldsymbol s}_n \rangle
\end{equation}
By virtue of Property~\ref{property:ExpansionKernels}, we can express the primal coefficients $a_k$ as 
\begin{equation}
a_k = \sum_{n=0}^N \eta_n s^k_n \Rightarrow
{\mathbf a} = \sum_{n=0}^N \eta_n {\boldsymbol s}_n
\end{equation}
where $\eta_n$ are the SVM Lagrange multipliers, and the solution at instant $m$ is
\begin{equation}
 {\hat y}_m = \sum_{n=0}^N \eta_n \langle {\boldsymbol s}_n,{\boldsymbol s}_m \rangle
\end{equation}
Only instants $n$ with $\eta_n\neq 0$ are part of the solution (Support Time Instants).
\end{theorem}

Therefore, each expansion coefficient $a_k$ can be expressed as a linear combination of input space vectors. Sparseness can be obtained in coefficients $a_k$, but not in coefficients $\eta_n$. Robustness is also ensured for coefficients $a_k$. The Lagrange multipliers are obtained from the dual problem, which is built in terms of a kernel matrix depending on the signal correlation.

\begin{definition}[Correlation Matrix from {Time-Transversal} Vectors]
Given the set of {time-transversal} vectors, the correlation matrix of the PSM is defined as 
\begin{equation} \label{eq:matcofprimal}
 \boldsymbol{R}^{\boldsymbol{s}}(m,n) \equiv \langle {\boldsymbol s}_m, {\boldsymbol s}_n \rangle = \sum_{k=0}^K s_m^{(k)} s_n^{(k)}
\end{equation}
\end{definition}

{In this setting, correlation matrix in Eq. \eqref{eq:matcofprimal} contains all the temporal correlations conveyed by the explanatory signals for different lags.} For instance, elements in the main diagonal will convey the zero-lag correlations between the {time-transversal} vectors for each time instant, whereas the upper and lower diagonal will convey the correlations between {time-transversal} vectors for lags $+1$ and $-1$. For the particular case of $s_n^{(k)}=s_{n-k}$, thus these signals being delayed versions of a signal $s_n$, the matrix is a Toeplitz matrix of the autocorrelation function of $s_n$. 
In summary, this correlation matrix consists of all the lags for the time correlations of the Hilbert signal subspace, which is always a fundamental information when working with time series.

\begin{property}[Correlation Matrix and Dual Problem]
Given the general PSM in~\eqref{eq:primalsignalmodel} and the correlation matrix from the {time-transversal} vectors in~\eqref{eq:matcofprimal}, the dual problem yielding the Lagrange multipliers consists of maximizing
\begin{equation} \label{eq:dualPSM}
 -\frac{1}{2} (\boldsymbol{\alpha} - \boldsymbol{\alpha}^\ast)^\top 
 	\left( \boldsymbol{R}^{\boldsymbol{s}} +\delta{\boldsymbol I} \right) 
 	(\boldsymbol{\alpha} - \boldsymbol{\alpha}^\ast) + (\boldsymbol{\alpha} - \boldsymbol{\alpha}^\ast)^\top \boldsymbol{y} -
 	\varepsilon \boldsymbol{1}^\top (\boldsymbol{\alpha} + \boldsymbol{\alpha}^\ast)
\end{equation}
constrained to $0 \leq \alpha_n,\alpha_n^\ast \leq C$.  
\end{property}

This property can be readily shown from considerations on the Lagrange functional and the associated KKT conditions~\cite{Rojo05a}. Therefore, by taking into account the PSM for a given DSP problem, one can determine the signals $s_n^{(k)}$ that generate the Hilbert subspace where the observations are projected to, and then the remaining elements and steps of the SVM methodology, {such as the input space, the input space correlation matrix, the dual QP problem, and the solution,} can be straightforwardly obtained.

\subsection{Spectral Analysis and System Identification}

The first {SVM algorithms for DSP} that were proposed using the PSM framework were the sinusoidal decomposition~\cite{Rojo03c}, the ARX system identification~\cite{Rojo04}, and the $\gamma$-filter structure~\cite{Camps04}. We next point out the relevant elements that can be identified in these algorithms.

\begin{property}(PSM Coefficients for Nonparametric Spectral Analysis).
Given the signal model hypothesis for nonparametric spectral analysis in Property~\ref{prop:sinusoidalModel}, estimated coefficients using the PSM are
\begin{equation} \label{eq:cyd}
 B_k = \sum_{n=0}^{N} \eta_n \cos(k \omega_0 t_n);
 \hspace{3mm}
 C_k = \sum_{n=0}^{N} \eta_n \sin(k \omega_0 t_n).
\end{equation}
\end{property}

\begin{property}(PSM Correlation and Dual Problem for Nonparametric Spectral Analysis). 
Given the signal model hypothesis in Property~\ref{prop:sinusoidalModel}, the correlation matrix is given by the sum of two terms,
\begin{eqnarray}
{\boldsymbol R}^{cos}(m,n) & = & \sum_{k=0}^{K} \cos(k \omega_0 t_m) \cos(k \omega_0 t_n)\\
{\boldsymbol R}^{sin}(m,n) & = & \sum_{k=0}^{K} \sin(k \omega_0 t_m) \sin(k \omega_0 t_n)
\end{eqnarray}
and the dual functional is given by~\eqref{eq:dualPSM} using ${\boldsymbol R}^{\boldsymbol s} =
{\boldsymbol R}^{cos} + {\boldsymbol R}^{sin}$.
\end{property}
The identification of the corresponding {time-transversal} vectors is straightforward. The derivation of this algorithm in~\cite{Rojo03c} is obtained by using these two properties in the PSM. Similar considerations can be drawn for ARX system identification algorithm in~\cite{Rojo04} using the next two properties.
\begin{property}[PSM Coefficients for ARX System Identification]
Given the signal model hypothesis for ARX system identification in Property~\ref{prop:ARXmodel},  estimated PSM coefficients are
\begin{equation}
 D_k = \sum_{n=0}^N \eta_n y_{n-k}; \hspace{.5cm}
 E_k = \sum_{n=0}^N \eta_n x_{n-k+1}.
\end{equation}
\end{property}
\begin{property}(PSM Correlation and Dual Problem for ARX System Identification).
Given the signal model hypothesis for ARX system identification in Property~\ref{prop:ARXmodel}, the correlation matrix is given by the sum of two terms,
\begin{eqnarray}
 \label{leafy} {\boldsymbol R}^y(m,n) & = & \sum_{k=1}^P y_{m-k}y_{n-k}, \\
 \label{leafx} {\boldsymbol R}^x(m,n) & = & \sum_{k=0}^Q x_{m-k+1}x_{n-k+1}.
\end{eqnarray}
These equations represent the time-local $P^{th}$ and $Q^{th}$ order sample estimators of the values of the (non-Toeplitz) autocorrelation functions of the input and the output discrete time
processes, respectively. The dual functional to be maximized is given by~\eqref{eq:dualPSM} using ${\boldsymbol R}^{\boldsymbol s} = {\boldsymbol R}^y + {\boldsymbol R}^x$.
\end{property}

\subsection{Convolutional Signal Models}

Convolutional signal model equations are those models that contain a convolutive mixture in their formulation. The most representative ones are the nonuniform interpolation (using {\em sinc} kernels, RBF kernels, or others) and the sparse deconvolution, presented in~\cite{Rojo07a,Rojo08a}. These models are relevant not only for their robustness, but also because their analysis gives us the foundations of the DSM to be subsequently used in a variety of {DSP} problem statements.
We next focus on summarizing the properties that are relevant for giving a signal processing block structure, that will be used for their analysis. 

\begin{property}[PSM Coefficients for Sinc Interpolation]
Given the signal model hypothesis in Property~\ref{prop:sincmodel} for {\em sinc} kernel interpolation, the PSM coefficients are
\begin{equation}	\label{eq:primalsinccoef}
 a_k = \sum_{n=0}^N \eta_n \textrm{sinc}(\sigma_0(t_k-t_n)).
\end{equation}
\end{property}
\begin{property}(PSM Correlation and Dual Problem for Sinc Interpolation).
Given the signal model hypothesis in Property~\ref{prop:sincmodel}, the correlation matrix is given by
\begin{equation}
{\boldsymbol R}^{sinc}(m,n)=\sum_{k=0}^N \textrm{sinc}(\sigma_0(t_m-t_k))\textrm{sinc}(\sigma_0(t_n-t_k))
\end{equation}
The maximized dual functional is in~\eqref{eq:dualPSM} when ${\boldsymbol R}^{\mathbf s} = {\boldsymbol R}^{sinc}$.
\end{property}

Coefficients in~\eqref{eq:primalsinccoef} are proportional to the cross correlation of coefficients $\eta_n$ and a set of sinc functions, each centered instants $t_n$~\cite{Rojo07a}.
Similar considerations can be made about the sparse deconvolution signal model equation.
\begin{property}[PSM Coefficients for Sparse Deconvolution]
The estimated PSM coefficients of the signal model hypothesis in Property~\ref{prop:sparsedeconmodel} are 
\begin{equation}\label{eq:sparse_model}
\hat x_n =\sum_{i=0}^{N} \eta_i h_{i-n} 
\end{equation}
\end{property}
\begin{property}(PSM Correlation and Dual Problem for Sparse Deconvolution).
The dual problem corresponding to Property~\ref{prop:sincmodel} is found by using the time-transversal vector 
\begin{equation}
{\boldsymbol s}^p = \left[ h_n, h_{n-1}, h_{n-2}, \dots, h_{n-{p+1}}\right]^\top.
\end{equation}
Correlation matrix ${\boldsymbol R^h}$ is given by~\eqref{eq:matcofprimal}, and in this case it represents the correlation matrix of ${\boldsymbol h}$. The dual functional to be maximized is given by~\eqref{eq:dualPSM}. 
\end{property}
In order to express the SVM algorithm for sparse deconvolution in terms of signal processing blocks, we can use Property~\ref{property:expansioncoefficients}, a well-known relationship between model residuals and Lagrange multipliers, valid for general SVM algorithms. This property establishes a non-linear relationship between the model coefficients $\eta_n$ and the residuals $e_n$ depending on the free parameters of the $\varepsilon$-Huber cost function.
 
Therefore, the Lagrange multipliers (or equivalently, the model coefficients) are mapped from the model residuals by using a static non-linear map which is given by the first derivative of the cost function (in our case, the $\varepsilon$-Huber cost). This property can be easily shown by using the appropriate set of Karush-Khun-Tucker conditions~\cite{Rojo05a}, and it indicates that the model coefficients are a piece-wise linear function of the model residuals. According to this expression of the Lagrange multipliers as a time series, the sparse deconvolution model can be further analyzed, as follows.
\begin{property}[PSM Block Diagram for Sparse Deconvolution] Let a discrete time signal be defined by model coefficients $\eta_n$ for $n = 0, \dots, N$, and being zero otherwise. Then, from eq. \eqref{eq:sparse_model} the relationship between the model coefficients and the estimated signal can be written as follows:
\begin{equation}\label{eq:convoletahache}
{\hat x}_n = \eta_n \ast h_{-n} \ast \delta_{n+M}
\end{equation}
where $\delta_n$ is the Kronecker delta sequence (1 for $n=0$ and zero elsewhere), the length of the impulse response $h_n$ is $M+1$ and $\ast$ denotes  discrete-time convolution operator.
\end{property}

\begin{figure}[t!] \centering
\IG[width=6cm]{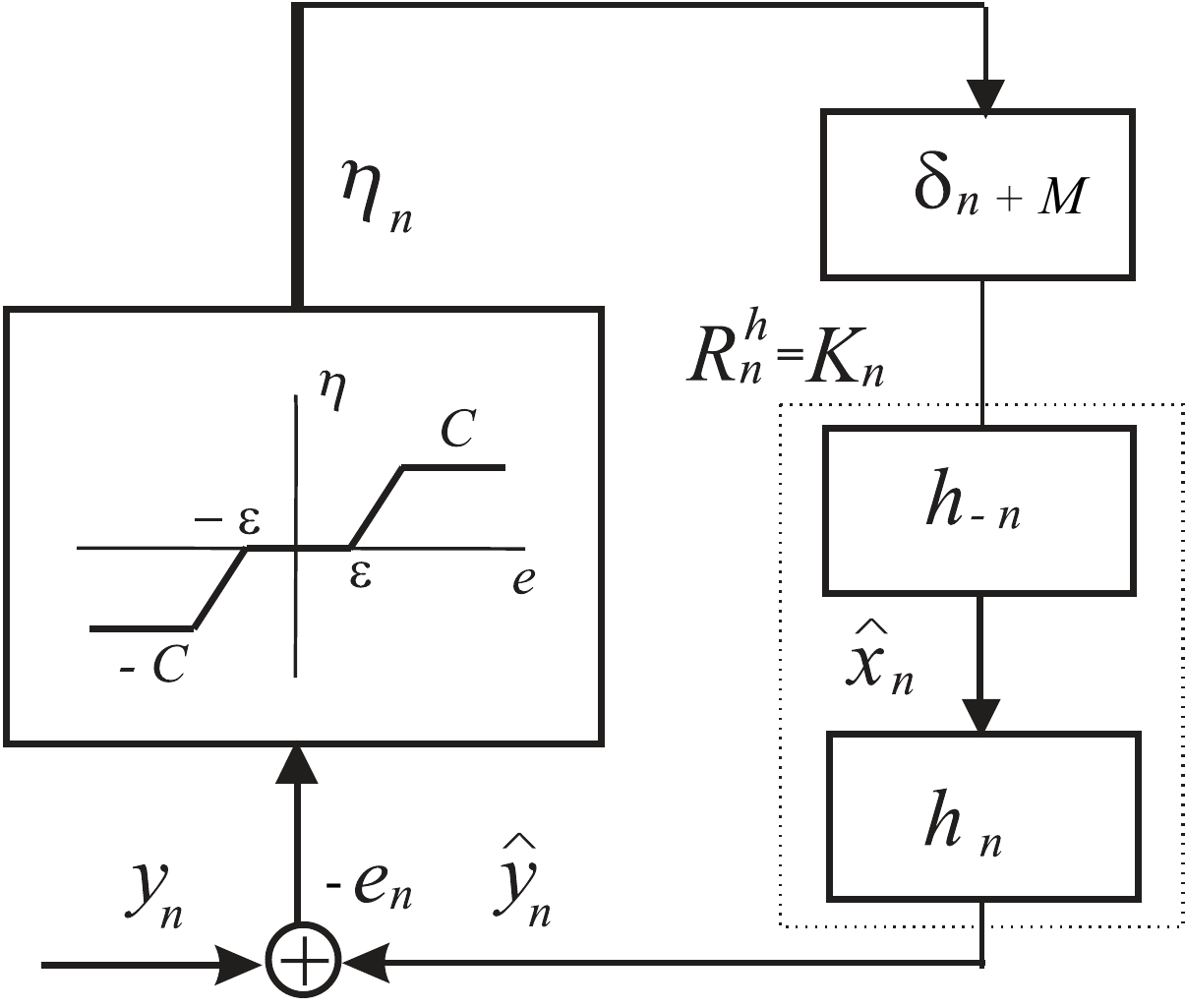}
\caption{Block diagram, elements and signals in the PSM Signal Model for SVM  sparse deconvolution.}\label{fig:decon_a}
\end{figure}

This is an interesting property from the signal processing point of view, and it can be easily obtained by examining the Karush-Khun-Tucker in the PSM of the sparse deconvolution problem~\cite{Rojo08a}. Hence, we can consider a joint equivalent closed-loop system, given in Fig.~\ref{fig:decon_a} which contains all the elements of the SVM algorithm expressed as signals or systems. Specifically, one is a non-linear system, given by Property~\ref{property:expansioncoefficients}, and the remaining ones are linear, time invariant systems. According to the preceding property, estimated signal ${\hat x}_n$ will not be sparse in general, because it is the sparseness of $\eta_n$ that can be controlled with the $\varepsilon$ parameter, but there is a convolutional relationship between ${\hat x}_n$ and $\eta_n$ that will depend on the impulse response, which in general does not have to be sparse. 

A particular class of kernels are {\it translation invariant kernels}, which are those fulfilling $K({\boldsymbol u},{\boldsymbol v}) = K({\boldsymbol u} - {\boldsymbol v})$. Two highly relevant properties in this setting, which will be useful in this section for PSM algorithms and later on for DSM algorithms, are the following.
\begin{property}[Shift-invariant Mercer's Kernels] \label{prop:shiftInvariant}
A necessary and sufficient condition for a translation invariant kernel to be Mercer's kernel
\cite{Zhang04} {is that its Fourier transform must be real and non-negative,} this is,
\begin{equation} \label{eq:translation}
\frac{1}{2 \pi}\int_{{\boldsymbol v} = -\infty}^{+\infty} K({\boldsymbol v})  e^{-j 2 \pi \langle {\boldsymbol f}, {\boldsymbol v}\rangle}
    d {\boldsymbol v} \geq 0 \hspace{3mm}\forall {\boldsymbol f} \in \mathbb{R}^d
\end{equation}
\end{property}
\begin{property} [Autocorrelation-Induced Kernel] \label{prop:autocorrelationKernel}
Let $\{h_n\}$ be a $(N+1)$-samples limited-duration discrete-time real signal, i.e., $\forall n \not \in (0,N) \Rightarrow h_n = 0$, and let $R_n^h = h_n * h_{-n}$ be its autocorrelation function. Then, the following shift-invariant
kernel can be built:
\begin{equation}
    K^h(n,m) = R_n^h(n-m)
\end{equation}
which is called Autocorrelation-Induced Kernel (or just autocorrelation kernel). As $R_n^h(m)$ is an even signal, its spectrum is real and nonnegative, and according to Property \ref{prop:shiftInvariant}, an autocorrelation kernel is always a Mercer's kernel.
\end{property}

Now, note that there is no Mercer's kernel appearing explicitly in the problem statement of PSM for sparse deconvolution, as it could be expected in {a SVM} approach. However, the block diagram  in Fig. \ref{fig:decon_a}  highlights that there is an implicitly present autocorrelation kernel, given by
\begin{equation}\label{eq:Rhn}
 R^h_n = h_n \ast h_{-n},
\end{equation}
in the case we associate the two systems containing the original system impulse response and its reversed version. From Linear System Theory, the order of the blocks could be changed without modifying the total system. However, the solution signal is embedded between these two blocks, which precludes the explicit use of this autocorrelation kernel in this PSM formulation. Finally, the role of delay system $\delta_{n+M}$ can be interpreted as just an index compensation that makes the total system causal.

In summary, the PSM algorithm yields a regularized solution, in which an autocorrelation kernel is implicitly used, but it does not allow to control the sparseness of the estimated signal. These properties will be used later in the DSM for high performance sparse deconvolution algorithms.

\subsection{Array Processing with Temporal Reference}

The array processing algorithm needs a complex-valued formulation. The complex Lagrange coefficients can be expressed as $\psi_n = \eta_n + j \nu_n$, $\eta_n=\alpha_n-\alpha_n^\ast$ and $\nu_n = \beta_n - \beta_n^\ast$ being the Lagrange coefficients generated by the real and imaginary parts of the error.
\begin{property}[PSM for Array Processing] Given the array processing signal defined in Property~\ref{prop:arraymodel}, the PSM coefficients for this problem are given by 
\begin{equation}
a_k=\sum_{n=0}^N \psi_n x^k_n
\end{equation}  
\end{property}
\begin{property}(Dual Problem for Temporal Reference Array Processing).\label{Prop:Complex}
Let ${y_n}$, with $0 \leq n \leq N$ be a set of desired signals available for training purposes, then the problem is known as temporal reference array processing. The incoming signal kernel matrix is defined as 
\begin{equation}
{\boldsymbol K}(l,m)=\sum_{i=0}^K {\overline x}^k_l x^k_m
\end{equation}
The dual functional to be maximized is a complex valued extension of~\eqref{eq:dualPSM}, i.e., 
\begin{equation}
{\boldsymbol \psi}^\top {\boldsymbol K}{\boldsymbol \psi}+\mathbb{R}e({\boldsymbol \psi}^\top {\boldsymbol {\overline {\boldsymbol y}}})-\varepsilon {\boldsymbol 1}^\top({\boldsymbol \alpha + {\boldsymbol\alpha}^\ast + {\boldsymbol\beta} + {\boldsymbol\beta}^\ast }),
\end{equation}
where $\overline {\boldsymbol y}$ stands for the complex conjugate of ${\boldsymbol y}$.
\end{property}

\subsection{PSM Application Examples}

The experiments in this section show several properties of the SVM elements and SVM algorithms presented up to this point. Specifically, the cost function in terms of robustness to outlying samples, the effect of the $\delta$ parameter, and the sparsity property are analyzed with the SVM algorithm for nonparametric spectral analysis~\cite{Rojo03c}.  
We also give an illustrative example on how to use the SVM-AR formulation for parametric spectral estimation. An additional example of PSM-DSP algorithms performance for {antenna array signal processing} is also included.

        \paragraph{PSM Algorithm for Nonparametric Spectral Analysis}

\begin{figure}[t!]
\begin{center}
\begin{tabular}{cc}
\IG[height=3cm]{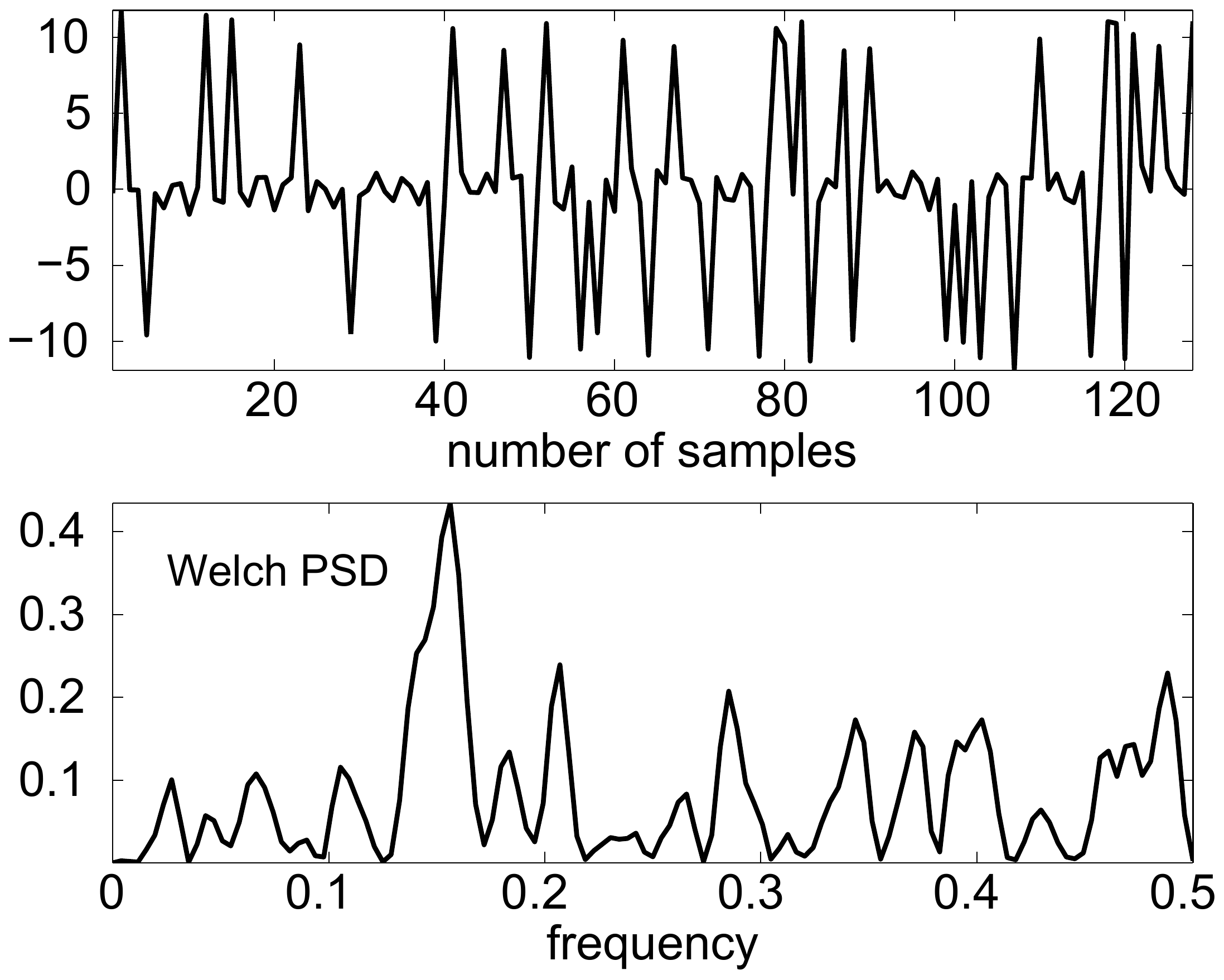} &
\IG[height=3cm]{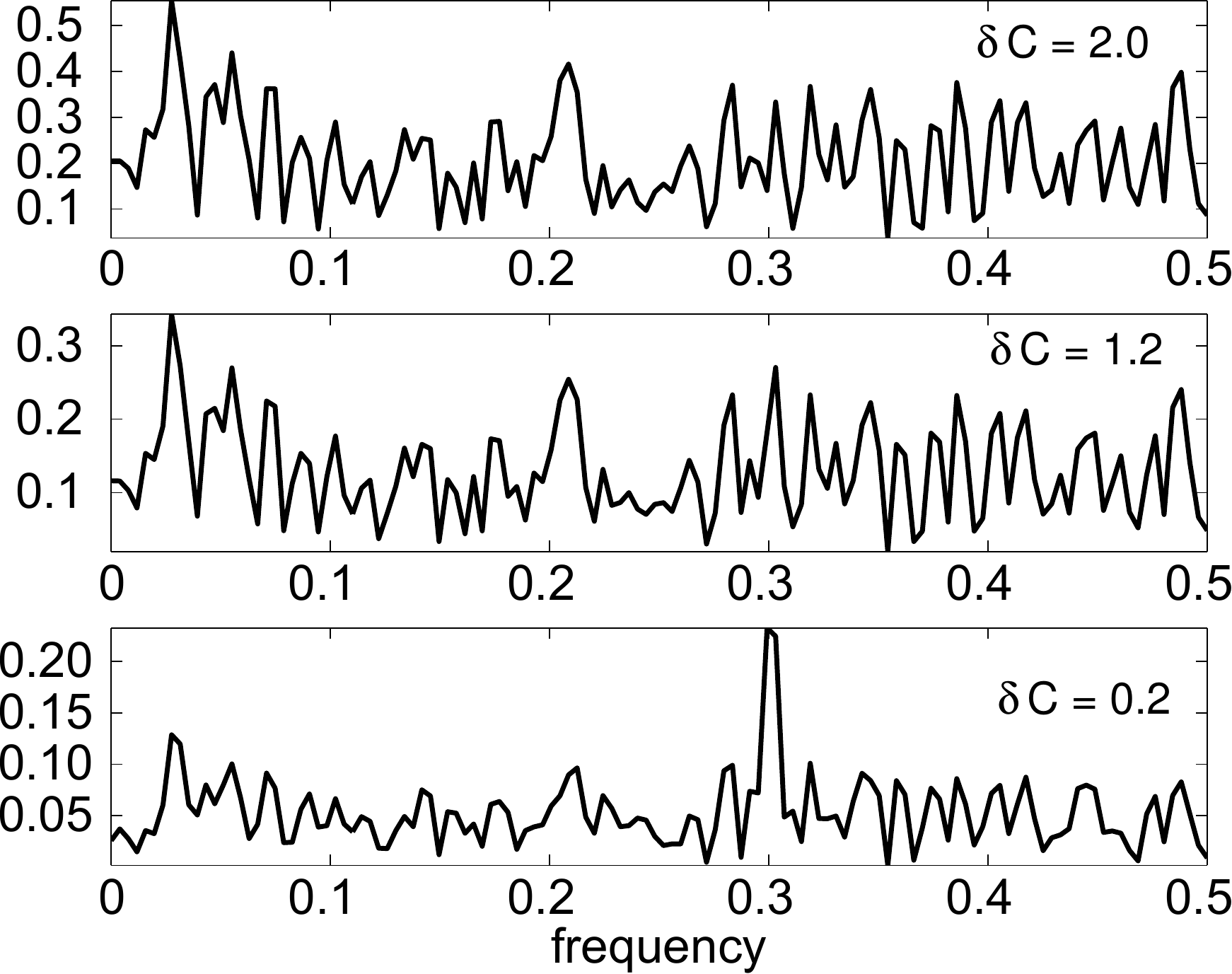}  \\
{\fns\it(a)} & {\fns\it(b)} \\
\multicolumn{2}{c}{\IG[height=6cm]{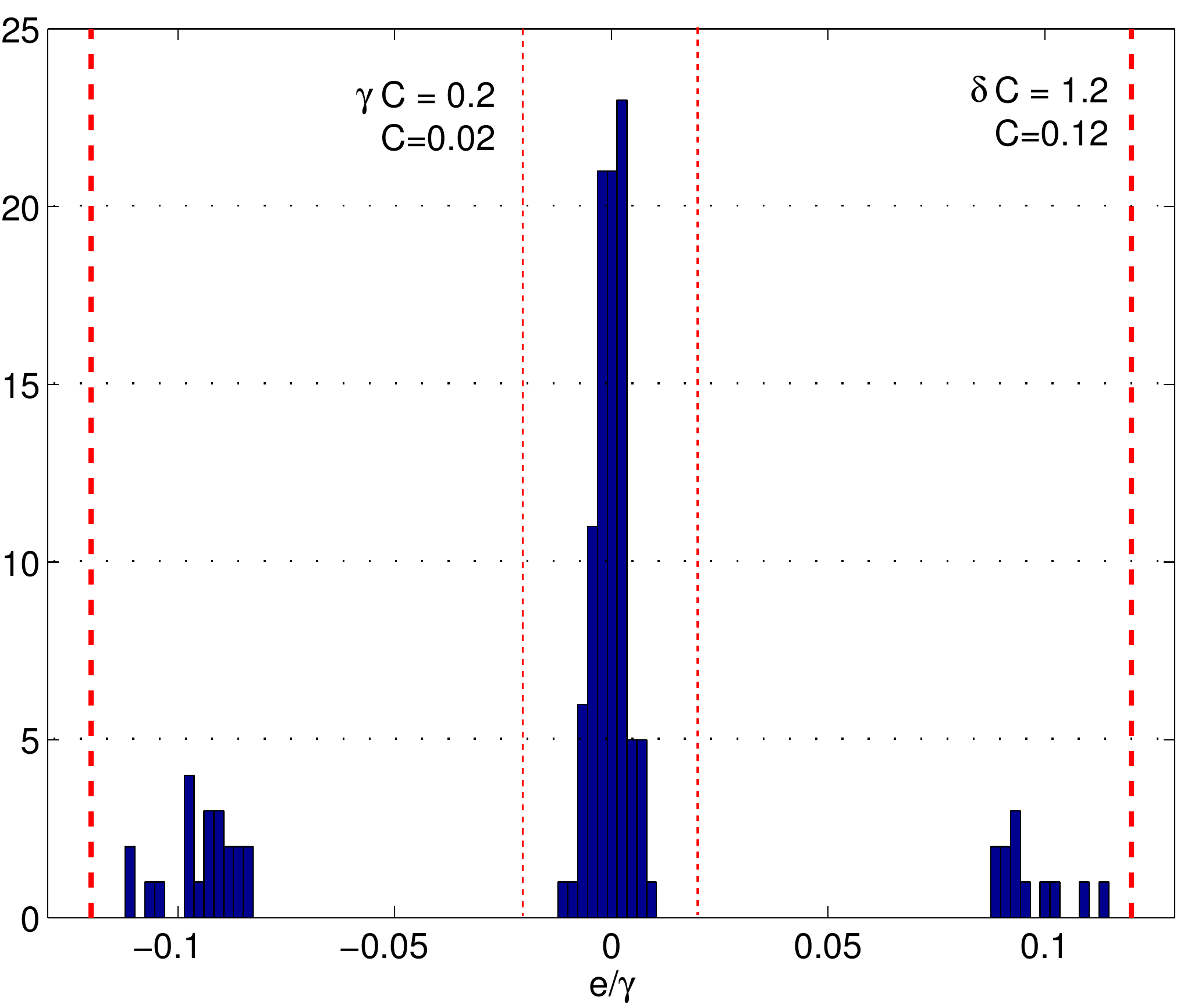}} \\ 
\multicolumn{2}{c}{\fns\it(c)} \\
\end{tabular}
\end{center}
\vspace{-0.65cm}
\caption{Insensitivity of PSM to impulse noise in nonparametric spectral analysis. (a) Sinusoid
 in impulsive noise (up) and its Welch periodogram (down). (b) SVM spectral estimators for different values of
 $\delta C$. (c) Histogram of the residuals (scaled to $\delta=10$) and
 control of the outlier impact on the solution with $C$. }
\label{fig:sinusoid}
\end{figure}

A synthetic data example shows the usefulness of PSM for dealing with outliers in the data. Let $y_n = \sin(2\pi f n) + e^v_n + e^j_n$, where $f=0.3$; $e^v_n$ is a white, Gaussian noise sequence (zero mean, variance 0.1); and $e^j_n$ is an impulsive noise process, generated as a sparse sequence for which 30\% of the randomly placed samples have high amplitude values given by $\pm10 + {\mathcal U}(-0.5,0.5)$, where ${\mathcal U}(\cdot)$ denotes the uniform distribution in the given interval, and the remaining are null samples. The length is $128$ samples, and we set $N_\omega=128$, see Fig.~\ref{fig:sinusoid}(a). We fixed $\varepsilon=0$ and $\delta=10$. Parameter $C$ can be chosen according to~\eqref{eq:KKTresid}. Figure~\ref{fig:sinusoid}(b) and (c) show that for $C$ low enough, large residual amplitudes can be present without impacting the solution. 

  \paragraph{Performance of PSM for AR Parametric Spectral Estimation}

\begin{figure}[t] \centering
\begin{tabular}{cc}
\IG[width=4.5cm]{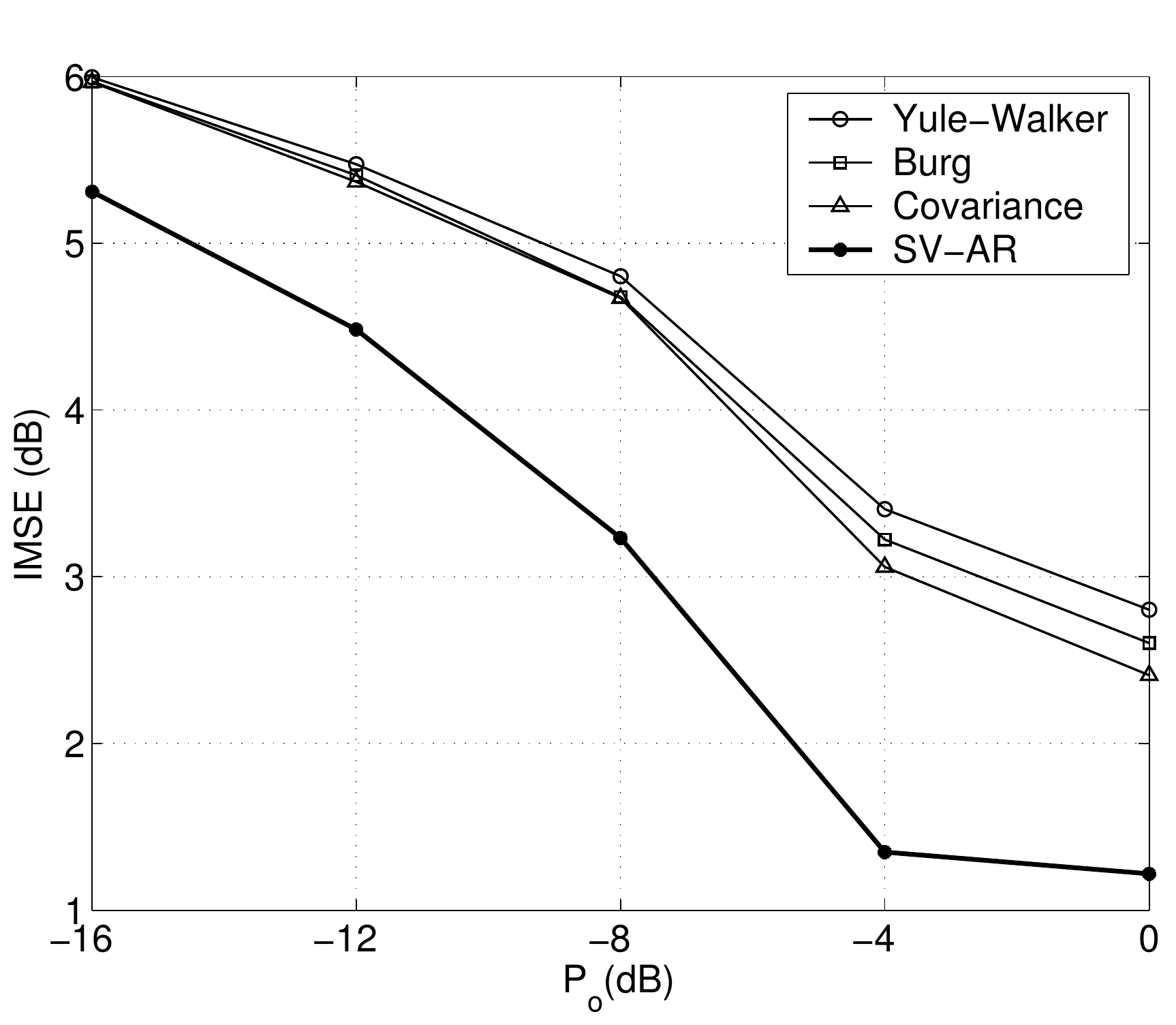} & \IG[width=4.5cm]{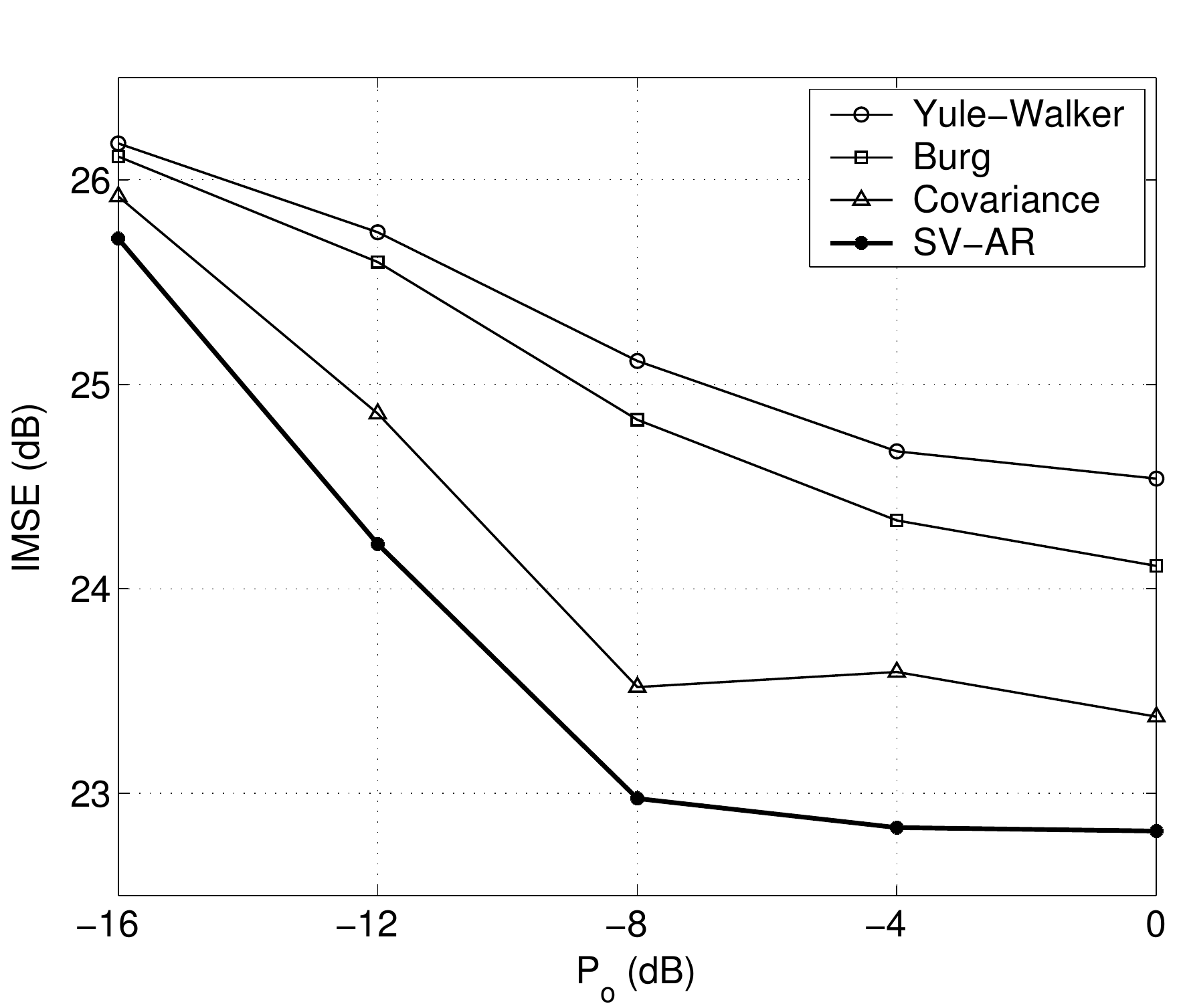} \\
{\fns\it(a)} & {\fns\it(b)} \\
\end{tabular}
\vspace{-0.5cm}
\caption{\it Application Example for Parametric Spectral Analysis. Evolution of the IMSE for different power of outliers for an (a) AR(3)-process and (b)
the ARMA(4,4) process.} \label{outliers}
\end{figure}
 
The most widely used linear system model for parametric spectral estimation is the all-pole structure~\cite{Marple87}. The output $y_n$ of such filter for a white noise input is an auto-regressive (AR) process of order $P$, AR($P$),
which can be expressed as $y_n = \sum_{p=1}^P D_p y_{n-p} + e_n $, 
where $D^p$ are the AR parameters, and $e_n$ denotes the samples of
the innovation process. Once the coefficients $D^p$ of the AR process
are calculated, the PSD estimation is
 $\Phi(f) = 
\sigma_P^2/f_s |1-\sum_{p=1}^{P} D_p e^{-j2\pi p
f/f_s}|^{-2}$,
where $f_s$ is the sampling frequency, and $\sigma_P^2$ is the
variance of the residuals.

We used data generated as an ARMA-process with $e_n$ given by white
Gaussian noise with zero mean and unit variance. Two systems,
previously introduced~\cite{Soderstrom87}, were analyzed, namely, an 
AR(3)-process:
$$y_n = e_n - 0.9816 y_{n-1} - 0.9400 y_{n-2} - 0.7799 y_{n-3}$$
and a narrow-band ARMA(4,4)-process:
$$y_n = e_n + 0.4800 e_{n-1} + 0.6876 e_{n-2} + 0.4476 e_{n-3} + 0.3538 e_{n-4} +$$ 
$$ + 1.0200 y_{n-1} - 2.0902 y_{n-2} + 0.9808 y_{n-3} - 0.9275 y_{n-4}.$$
The input discrete process was a ${\mathcal N}(0,1)$ sequence with $L=128$ samples length.
The output signal was corrupted by additive noise ${\mathcal N}(0,0.1)$, and 20\% of samples were affected by
impulsive noise from a zero mean and unit variance uniform distribution and
randomly placed. These $L$ samples were used for training the model and 1000 samples more were used
for validation. For all simulations, parameters $C$ and $\delta$ were searched in the
range $[10^{-5},10^{5}]$, and we fixed $\varepsilon=0$. The performance criterion used for the general estimate of
$\Phi(f)$ was the Integrated Mean-square Error (IMSE), given by 
$\text{IMSE} = \frac{1}{N_F} \sum_{f=1}^{N_F} \big| \Phi(f) - \hat \Phi(f)\big|^2$,
where $N_F$ is the number of estimated frequencies in the spectrum. The experiment was repeated 100
times and the best model was selected according to the estimated IMSE in the validation set. Figure~\ref{outliers} illustrates the effect of different power of outliers $P_o$ on the estimation accuracy. In both systems, the SVM-AR method outperformed the standard methods in all situations, with an average gain of 1.5 to 2 dB. Differences between the methods are lower with increasing noise, specially for the ARMA(4,4).

  \paragraph{SVM  Array Processing with Temporal Reference}

A linear array of six elements spaced $d=0.51 \lambda$, was used to detect the signal from a desired user in presence of different interferences in an environment with Gaussian noise~\cite{Martinez05}. The desired signal was assumed to experiment small multi-path propagation coming from DOAs $-0.1\pi$ and $-0.25 \pi$, with amplitudes $1$ and $0.3$, respectively, and different phases. Two interferences signals came from $-0.05 \pi$, $0.1 \pi$, and $0.3 \pi$, all of them with unit amplitude and phase $0$. The desired signal was organized in bursts of 50 training samples plus $10^5$ test samples. 

The SVM was compared to the standard LS algorithm for array processing. Since the noise was assumed to be thermal, then its variance could be assumed to be approximately known. In communications, parameter validation is usually not affordable due the small amount of available data and the low computational power of systems. Therefore, parameter $\delta $ of the SVM cost function was set to $10^{-6}$, and $C$ was set in order to adjust $\delta C$ to the thermal noise standard deviation. Hence, residuals for samples corrupted mainly by thermal noise were likely in the quadratic cost, and residuals for samples with high error (when interfering signals added in phase) were likely in the linear cost.

\begin{figure}[h!]
\centering
\includegraphics[width=6cm]{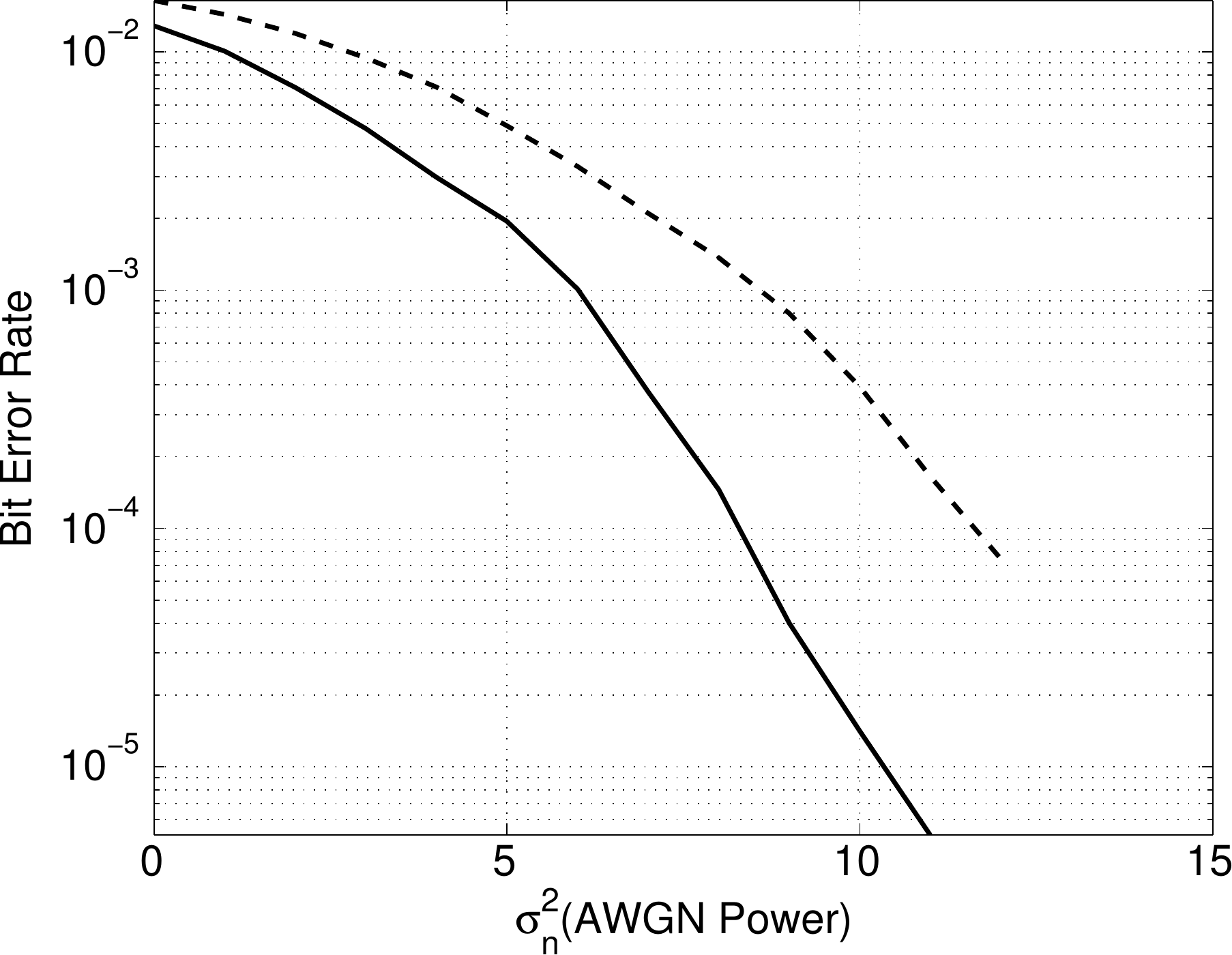}
\vspace{-0.3cm}
\caption{Application example for Array Processing with Temporal Reference. {BER performance of the SVM (solid)} and regular LS (dashed) beamformers, in the presence of thermal noise plus interfering signals giving large residuals.}
\label{fig:BER_SVM_Linear}
\end{figure}

Results are depicted in Fig.~\ref{fig:BER_SVM_Linear}. Each Bit Error Ratio (BER) was evaluated by averaging 100 independent trials. The LS criterion is highly biased by the non-Gaussian nature of the data produced by the multipath environment plus the interfering signals. SVM is closer to the linear optimal and offers a processing gain of several dBs {with respect to the LS solution.}

\section{Type II Algorithms: RKHS Signal Models}		\label{rkHs}

The class of {\it RSM algorithms} algorithms consists of stating the signal model equation of the time series structure in the RKHS. The major interest of this approach is the combination of flexibility (provided to the algorithm by the non-linearity) together with the possibility of scrutinizing the time series structure of the model and the solution. {It can be seen in Tables \ref{tab:scheme1} and \ref{tab:scheme2} that signal model equations are stated in the RKHS in this kind of algorithms, which appears to be evident in the ARX case example.}

\subsection{Fundamentals of RSM}

In this section, we state a general signal model equation for estimation with discrete-time series notation. Nonlinear ARX system identification use a signal model equation in the RKHS relating an exogenous time series and an observed data time series, whereas antenna array processing with spatial reference uses an energy expression in the RKHS, together with complex-valued algebra. 

\begin{theorem}[RSM Problem Statement]
Let $\{y_n\}$ be a discrete-time series and ${\boldsymbol v}_n$ a vector observation for each time instant.
A non-linear signal model equation can be stated by transforming the weight vector and the input vectors at time instant $n$ to an RKHS,
\begin{equation}\label{eq:rkHsmodel}
 {\hat y}_n = \langle {\boldsymbol w},{\boldsymbol \varphi}({\boldsymbol v}_n) \rangle + b
\end{equation}
Since the same signal model equation is used with weight vector ${\boldsymbol w} = \sum_{n=0}^N \eta_n {\boldsymbol \varphi}({\boldsymbol v}_n)$, the solution can be expressed as
\begin{equation}\label{eq:rkHsdualmodel}
 {\hat y}_m = \sum_{n=0}^N \eta_n \langle {\boldsymbol \varphi}({\boldsymbol v}_n),{\boldsymbol \varphi}({\boldsymbol v}_m) \rangle = \sum_{n=0}^N \eta_n K( {\boldsymbol v}_n,{\boldsymbol v}_m ) + b
\end{equation}
\end{theorem}
The proof is given by using the conventional Lagrangian functional and the dual problem. This is the most used approach to state data problems with SVM. This theorem is next used to obtain the non-linear equations for several DSP problems.
Composite kernels are used for defining relationships between two observed (exogenous and output) signals, by means of an RKHS system identification model.
\begin{property}[Composite Summation Kernel]
A simple composite kernel comes from the concatenation of non-linear transformations of ${\boldsymbol c} \in \mathbb{R}^c$ and ${\boldsymbol d} \in \mathbb{R}^d$, given by 
\begin{equation}
{\boldsymbol \phi}({\boldsymbol c},{\boldsymbol d})=\{{\boldsymbol \phi}_1({\boldsymbol c}),{\boldsymbol \phi}_2({\boldsymbol d})\}
\end{equation}
where $\{\cdot,\cdot\}$ denotes concatenation of column vectors, and ${\boldsymbol \phi}_1(\cdot)$, ${\boldsymbol \phi}_2(\cdot)$ are transformations to Hilbert spaces ${{\mathcal H}}_1$ and ${{\mathcal H}}_2$. The dot product between vectors is
\begin{equation}
K({\boldsymbol c}_1,{\boldsymbol d}_1;{\boldsymbol c}_2,{\boldsymbol d}_2) =
 \langle {\boldsymbol \phi}({\boldsymbol c}_1,{\boldsymbol d}_1), {\boldsymbol \phi}({\boldsymbol c}_2,{\boldsymbol d}_2)\rangle
 = K_1({\boldsymbol c}_1,{\boldsymbol c}_2) + K_2({\boldsymbol d}_1,{\boldsymbol d}_2)
\end{equation}
which is known as summation kernel. 
\end{property}
The composite kernel expression of a summation kernel can also account for the cross information between an exogenous and an output observed time series. 
\begin{property}[Composite kernels for Cross Information] \label{prop:compositecross}
Assume a non-linear mapping ${\boldsymbol \varphi}(\cdot)$ into a Hilbert space ${{\mathcal H}}$ and three linear transformations ${\boldsymbol A}_i$ from ${{\mathcal H}}$ to ${{\mathcal H}}_i$, for $i=1,2,3$. We construct the following composite vector,
\begin{equation}
{\boldsymbol \phi}({\boldsymbol c},{\boldsymbol d})=\{{\boldsymbol A}_1 {\boldsymbol \varphi}({\boldsymbol c}),{\boldsymbol A}_2
{\boldsymbol \varphi}({\boldsymbol d}),{\boldsymbol A}_3 ({\boldsymbol \varphi}({\boldsymbol c})+{\boldsymbol \varphi}({\boldsymbol d}))\}
\end{equation}
If the dot product is computed, we obtain
\begin{equation}
\begin{split}\label{4k}
K({\boldsymbol c}_1&,{\boldsymbol c}_2;{\boldsymbol d}_1,{\boldsymbol d}_2)  = \\
& {\boldsymbol \phi}^\top({\boldsymbol c}_1) {\boldsymbol R}_1 \phi({\boldsymbol c}_2)+
  {\boldsymbol \phi}^\top({\boldsymbol c}_1){\bf R}_2\phi({\boldsymbol d}_2) \\
&~~~~~~~~~~~+{\boldsymbol \phi}^\top({\boldsymbol d}_1){\bf R}_3\phi({\boldsymbol c}_2)+ 
  {\boldsymbol \phi}^\top({\boldsymbol d}_1) {\bf R}_3\phi({\boldsymbol c}_2) \\
& = K_1({\boldsymbol c}_1,{\boldsymbol c}_2) + K_2({\boldsymbol d}_1,{\boldsymbol d}_2) + K_3({\boldsymbol c}_1,{\boldsymbol d}_2) + K_3({\boldsymbol d}_1,{\boldsymbol c}_2) \nonumber
\end{split}
\end{equation}
where ${\boldsymbol R}_1={\boldsymbol A}^\top_1{\bf A}_1+{\bf A}^\top_3{\bf A}_3$, ${\bf R}_2={\bf A}^\top_2{\bf A}_2+{\bf A}^\top_3{\bf A}_3$ and ${\bf R}_3={\bf A}^\top_3{\bf A}_3$ are three independent definite positive matrices.
\end{property}
Note that, in this case, ${\boldsymbol c}$ and ${\boldsymbol d}$ must have the same dimension.

\subsection{Nonlinear ARX Identification}

In~\cite{Drezet98, Gretton01, Suykens01, Goethals05,Espinoza05}, the SVR algorithm was used for non-linear system identification, but the time series structure of the data was not scrutinized. In~\cite{Rojo04}, SVM was explicitly formulated for modeling linear time-invariant ARMA systems (linear SVM-ARMA), and it was extended to a general framework for linear signal processing problems~\cite{Rojo05a}. When linearity cannot be assumed, non-linear system identification techniques are required. {General non-linear models, such as artificial neural networks, wavelets, and fuzzy models, are common and effective choices~\cite{Ljung99,Nelles01}, but the temporal structure cannot be easily analyzed as far as it remains inside a black-box model. This is the main problem for standard non-linear approaches used in the NARX approach.}

We next summarize several SVM procedures for non-linear system identification {that has alleviated these problems.} The material has been presented in detail in~\cite{Martinez06a,CampsValls07,Camps08a}, so we focus here on the most relevant aspects. First, the stacked SVR algorithm for non-linear system identification is briefly examined in order to check that, though efficient, this approach does not correspond explicitly to an ARX model in the RKHS. 

Let $\{x_n\}$ and $\{y_n\}$ be two discrete-time signals, which are the input and the output, respectively, of a non-linear system. Let ${\boldsymbol y}_n = [y_{n-1},y_{n-2},\dots,y_{n-M}]^\top$ and ${\boldsymbol x}_n = [x_n,x_{n-1},\dots,x_{n-Q+1}]^\top$ denote the states of input and output at time instant $n$.
The stacked-kernel system identification algorithm~\cite{Gretton01,Suykens01} is next described. 

\begin{property}(Stacked-kernel Signal Model for SVM  Nonlinear System Identification).
 Assuming a non-linear transformation ${\boldsymbol \phi}(\{{\boldsymbol y}_n,{\boldsymbol x}_n\})$ for the concatenation of the input and output discrete time processes to a B-dimensional feature space, $\phi: \mathbb{R}^{M+Q}\rightarrow \cal{H}$, a linear regression model can be built in $\cal{H}$, given by
\begin{equation}\label{differenceEquationRegress}
 y_n =\langle {\boldsymbol w},{\boldsymbol \phi}(\{{\boldsymbol y}_n,{\boldsymbol x}_n\}) \rangle + e_n,
\end{equation}
where ${\boldsymbol w}$ is a vector of coefficients in the RKHS, given by
\begin{equation}
 {\boldsymbol w} = \sum_{n=0}^N \eta_n {\boldsymbol \phi}(\{{\boldsymbol y}_n,{\boldsymbol x}_n\}), \label{predictd}
\end{equation}
and the following Gram matrix containing the dot products can be identified
\begin{equation}\label{eq:Gram}
{\boldsymbol G} (m,n) = \langle {\boldsymbol \phi}(\{{\boldsymbol y}_m,{\boldsymbol x}_m\}), 
{\boldsymbol \phi}(\{{\boldsymbol y}_n,{\boldsymbol x}_n\}) 
= K(\{{\boldsymbol y}_m,{\boldsymbol x}_m\},\{{\boldsymbol y}_n,{\boldsymbol x}_n\}).
\end{equation}
The non-linear mappings do not need to be explicitly computed, but instead the dot product in RKHS can be replaced by Mercer's kernels. Then, the predicted output for newly observed $\{{\boldsymbol y}_m,{\boldsymbol x}_m\}$ is given by
\begin{equation}\label{eq:solutionSVRident}
 \hat{y}_m = \sum_{n=0}^N \eta_n K(\{{\boldsymbol y}_m,{\boldsymbol x}_m\},\{{\boldsymbol y}_n,{\boldsymbol x}_n\}).
\end{equation}
\end{property}
Note that this is the expression for a general non-linear system identification, though it does not correspond to an ARX structure in the RKHS. Although the reported performance of the algorithm is high when compared to other approaches, this formulation does not allow us to scrutinize the statistical properties of the time series that are being modeled in terms of autocorrelation and/or cross correlation between the input and the output time series. Composite kernels can be introduced, allowing us to create a non-linear version of the linear SVM-ARX algorithm by actually using an ARX scheme on the RKHS signal model. 

 \begin{figure}[t]
 \centerline{\IG[width=4cm]{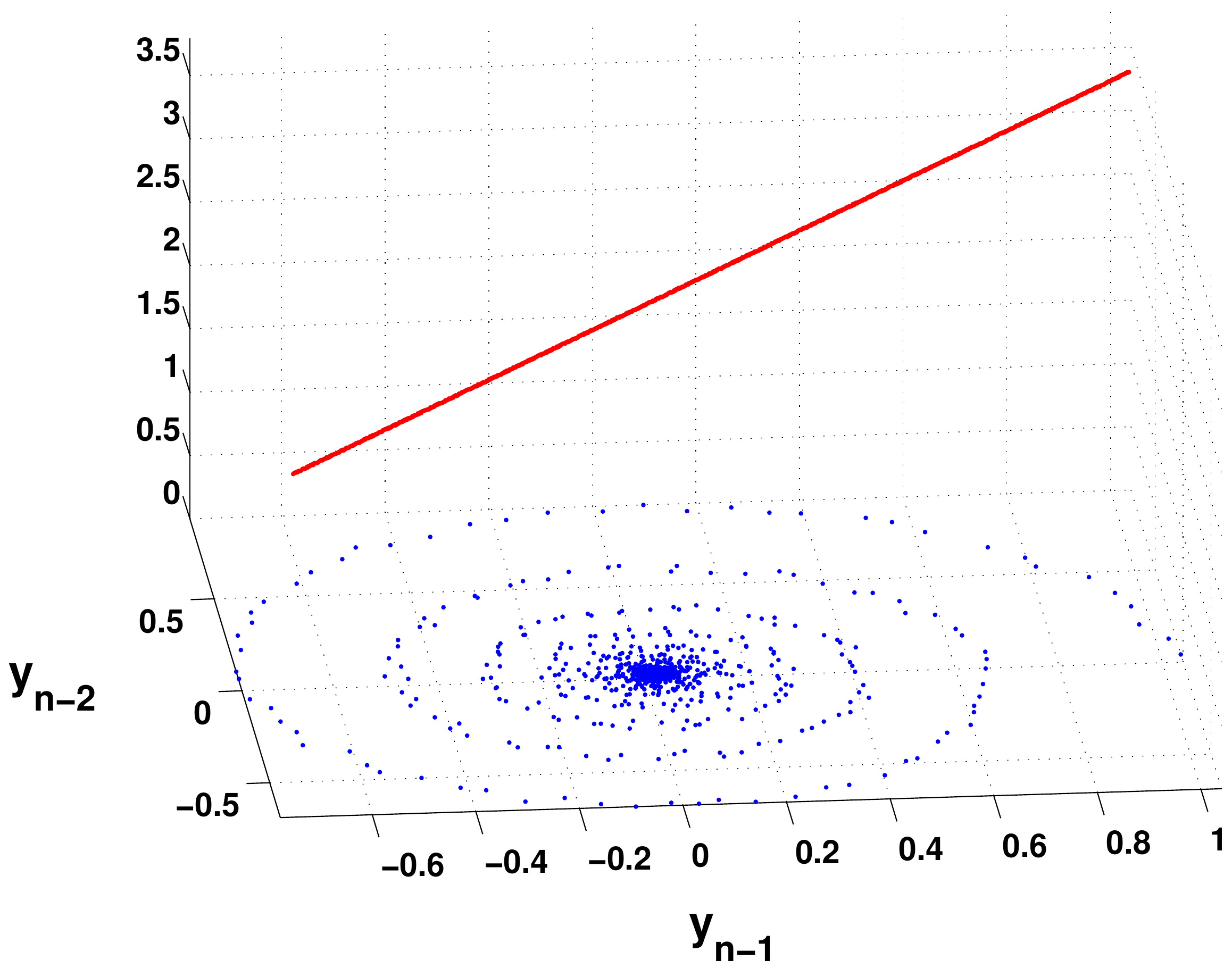} \IG[width=4cm]{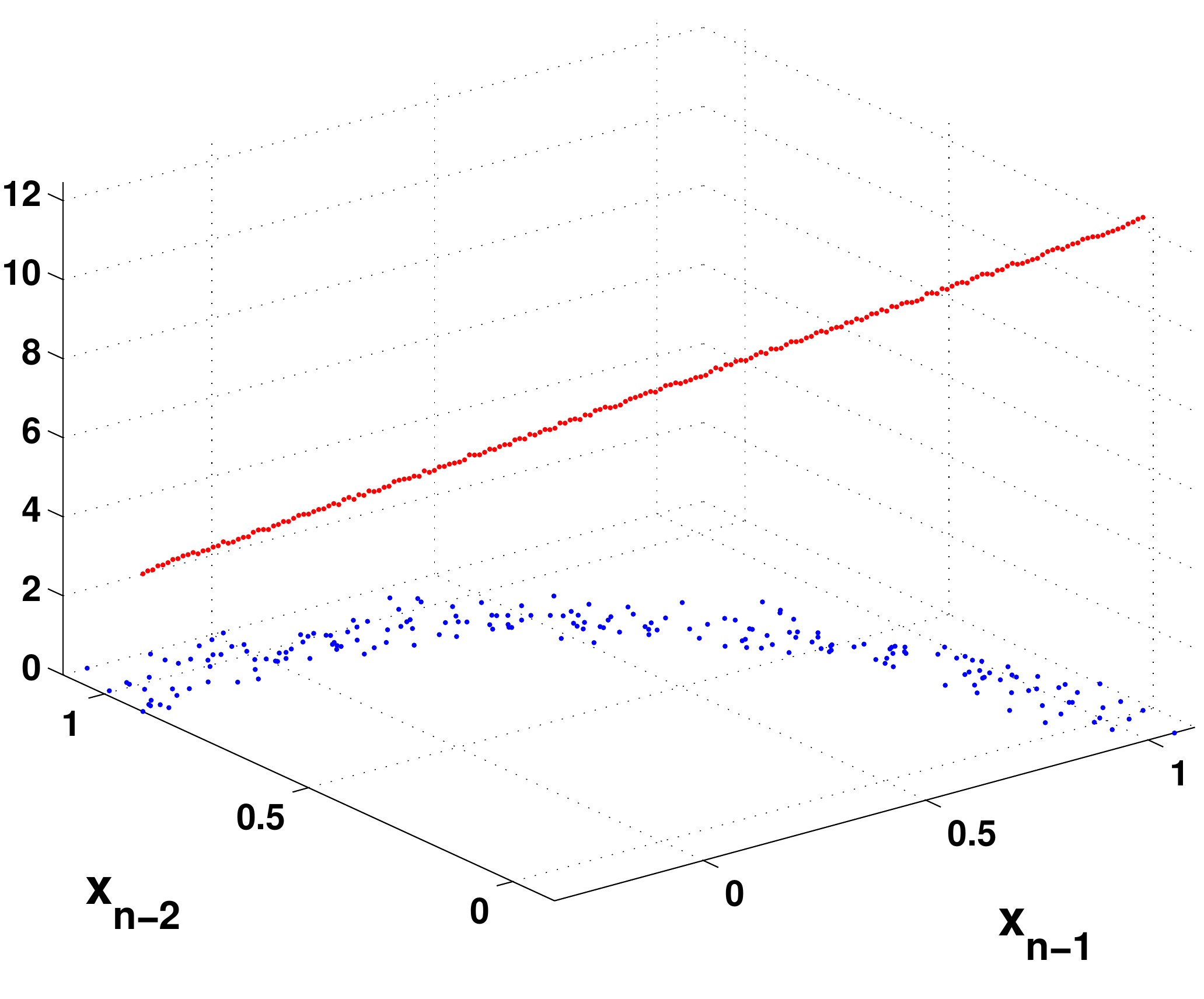}}
 \vspace{-0.3cm}
 \caption{{A non-linear relationship between samples in the input space is transformed into a linear relationship in the high dimensional RKHS.} Different signal model equations can be used in the RKHS, for instance, the use of different kernels for the AR and the MA components of the difference equation in a ARX system identification problem allows to implement different non-linearities for the input and output of the non-linear system.}
 \label{fig:rkHs_mapping}
 \end{figure}

\begin{property}(SVM  in RKHS for ARX Nonlinear System Identification). 
If we separately map the state vectors of both the input and the output discrete-time signals to ${\mathcal H}$, using a non-linear mapping given by ${\boldsymbol \phi}_e({\boldsymbol x}_n): \mathbb{R}^{M}\rightarrow {\mathcal H}_e$ and ${\boldsymbol \phi}_d({\boldsymbol
y}_n): \mathbb{R}^Q\rightarrow {\mathcal H}_d$, then a linear ARX model can be stated in ${\mathcal H}$, whose corresponding difference equation is given by
\begin{equation}\label{eq:diffARMA}
 y_n = \langle {\boldsymbol w}_d, {\boldsymbol \phi}_d({\boldsymbol y}_{n})\rangle 
 + \langle {\boldsymbol w}_e, {\boldsymbol \phi}_e({\boldsymbol x}_{n}) \rangle + e_n,
\end{equation}
where ${\boldsymbol w}_d$ and ${\boldsymbol w}_e$ are vectors determining the AR and the MA
coefficients of the system, respectively, in (possibly different) RKHS. The vector coefficients are
\begin{equation}
 {\boldsymbol w}_d = \sum_{n=0}^N \eta_n {\boldsymbol \phi}_d({\boldsymbol y}_{n}); \hspace{8mm}
 {\boldsymbol w}_e = \sum_{n=0}^N \eta_n {\boldsymbol \phi}_e({\boldsymbol x}_{n}),
\end{equation}
and two different kernel matrices can be further identified:
\begin{eqnarray} \label{leafyrkHs}
 {\boldsymbol R}^y(m,n) & = & \langle {\boldsymbol \phi}_d({\boldsymbol y}_{m}),{\boldsymbol \phi}_d({\boldsymbol y}_{n}) \rangle =
  K_d({\boldsymbol y}_{m}, {\boldsymbol y}_{n})\\ 
 {\boldsymbol R}^x(m,n) & = & \langle {\boldsymbol \phi}_e({\boldsymbol x}_{m}), {\boldsymbol \phi}_e({\boldsymbol x}_{k}) \rangle =
  K_e({\boldsymbol x}_{m}, {\boldsymbol x}_{n}).
\end{eqnarray}
These equations account for the sample estimators of both the input and
output time series autocorrelation functions~\cite{Papoulis91} in the RKHS. Actually, they are proportional 
to the non-Toeplitz estimator of the time series autocorrelation matrices.

The dual problem consists of maximizing~\eqref{eq:dualPSM} with ${\boldsymbol R^s} = {\boldsymbol R}^x+{\boldsymbol R}^y$, and the output for a new observation vector is obtained as
\begin{equation}\label{eq:solutionARMA}
\hat y_m = \sum_{n=1}^N \eta_n \left( K_d({\boldsymbol y}_n,{\boldsymbol y}_m)
 + K_e({\boldsymbol x}_n,{\boldsymbol x}_m) \right)
\end{equation}
\end{property}
The kernels in the preceding equation correspond to correlation matrices computed into the direct summation of kernel spaces $\mathcal{H}_1$ and $\mathcal{H}_2$. Hence, the autocorrelation matrices components given by ${\boldsymbol x}_n$ and ${\boldsymbol y}_n$ are expressed in their corresponding RKHS and the cross correlation component is computed in the direct summation space. A third space can be used to compute the cross correlation component, which introduces generality to the model. Figure \ref{fig:rkHs_mapping} shows a pictorial representation of the different kernels corresponding to different non-linearities which cannot be accommodated, for instance, by the use of a single width using a RBF kernel, but instead, they can be readily approximated by using two separate kernels for each time series or time process.

\begin{property}(Composite Kernels for General Cross Information in System Identification). 
Assuming a non-linear mapping $\varphi(\cdot)$ into
a Hilbert space ${{\mathcal H}}$ and three linear transformations ${\boldsymbol
A}_i$ from ${{\mathcal H}}$ to ${{\mathcal H}}_i$, for $i=1,2,3$, we can construct
the following composite vector:
\begin{equation}
\phi({\boldsymbol y},{\boldsymbol x})=\{{\boldsymbol A}_1 \varphi({\boldsymbol x}),{\boldsymbol A}_2
\varphi({\boldsymbol y}),{\boldsymbol A}_3 (\varphi({\boldsymbol x})+\varphi({\boldsymbol y}))\}
\end{equation}
According to Property~\ref{prop:compositecross}, we have
\begin{equation}
\begin{split}\label{4kbis}
K({\boldsymbol y}_m&,{\boldsymbol y}_n;{\boldsymbol x}_m,{\boldsymbol x}_n)  = \\
& {\boldsymbol \phi}^\top({\boldsymbol y}_m) {\boldsymbol R}_1 \phi({\boldsymbol y}_n)+
  {\boldsymbol \phi}^\top({\boldsymbol y}_m){\bf R}_2\phi({\boldsymbol x}_n) \\
&~~~~~~~~~~~+{\boldsymbol \phi}^\top({\boldsymbol x}_m){\bf R}_3\phi({\boldsymbol y}_n)+ 
  {\boldsymbol \phi}^\top({\boldsymbol x}_m) {\bf R}_3\phi({\boldsymbol y}_n) \\
& = K_1({\boldsymbol y}_m,{\boldsymbol y}_n) + K_2({\boldsymbol x}_m,{\boldsymbol x}_n) + K_3({\boldsymbol y}_m,{\boldsymbol x}_n) + K_3({\boldsymbol x}_m,{\boldsymbol y}_n) \nonumber
\end{split}
\end{equation}
where it is straightforward to identify $K_1 = K_d$ and $K_2 = K_e$.
\end{property}
Note that in this case, ${\boldsymbol x}_n$ and ${\boldsymbol y}_n$ need to have the same dimension, which can be naively accomplished by zero completion of the embeddings.
\begin{property}[General Composite Kernels]
A general composite kernel, that can be obtained as a combination of the previous ones, is given by
\begin{equation}
\begin{split}\label{4ksvr}
& K({\boldsymbol x}_m,{\boldsymbol y}_m;{\boldsymbol y}_n,{\boldsymbol x}_n) = 
K_1({\boldsymbol y}_m,{\boldsymbol y}_n)+K_2({\boldsymbol x}_m,{\boldsymbol x}_n) \\
& ~~~~~~~~~~~~+ K_3({\boldsymbol y}_{m},{\boldsymbol x}_n) + K_3({\boldsymbol x}_{m},{\boldsymbol y}_n) + K_4({\boldsymbol z}_m,{\boldsymbol z}_n)
\end{split}
\end{equation}
\end{property}
Therefore, despite the fact of SVM-ARX and SVR non-linear system identification are different problem statements, both models can be easily combined.

\subsection{Array Processing with Spatial Reference}

The array processing problem stated in~\eqref{eq:array_problem} can be solved when there are no training symbols available, but just a set of incoming data and information about the angle of arrival of the desired user. In this case, the algorithm to be applied consists of a processor that detects without distortion (distortionless property) the signal from the desired direction of arrival while minimizing the total output energy. The signal can be easily mapped to an RKHS, and then we minimize
\begin{equation}
\mathcal{E}=E({\boldsymbol w}^H\varphi({\boldsymbol x}_n)\varphi({\boldsymbol x}_n)^H{\boldsymbol w})={\boldsymbol w}^H{\boldsymbol R}{\boldsymbol w}\approx {\boldsymbol w}^H {\boldsymbol \varPhi}{\boldsymbol \varPhi}^H {\boldsymbol w}
\end{equation}
for a given set of previously collected snapshots, where $E$ stands for statistical expectation, and ${\boldsymbol \varPhi}$ is a matrix containing all mapped snapshots $\varphi({\boldsymbol x}_n)$.

\begin{property}[Spatial Reference Signal Model in RKHS]
In order to introduce the distortionless property, constraints must be applied to a set of canonical signals (spatial reference signals) whose steering vector~\eqref{eq:steering_vector} contains the desired direction of arrival $\theta_0$, carrying a set of symbols $b_i$. The reference signal model equation is
\begin{equation}\label{eq:power_min}
b_i={\boldsymbol w}^H \varphi(b_i \boldsymbol a_0)-b
\end{equation}
where ${\bf a}_0$ is the steering vector corresponding to the desired signal. 
Then, a primal functional must contain the following constraints
\begin{equation}\label{eq:constraints_reg2}
\begin{split}
\mathbb{R}e\left(b_i-{\boldsymbol w}^H \varphi(b_i \boldsymbol a_0)-b\right)&\leq \varepsilon + \xi_i\\
-\mathbb{R}e\left(b_i-{\boldsymbol w}^H \varphi(b_i \boldsymbol a_0)-b\right)&\leq \varepsilon + \xi'_i\\
\mathbb{I}m\left(b_i-{\boldsymbol w}^H \varphi(b_i \boldsymbol a_0)-b\right)&\leq \varepsilon + \zeta_i\\
-\mathbb{I}m\left(b_i-{\boldsymbol w}^H \varphi(b_i \boldsymbol a_0)-b\right)&\leq \varepsilon +\zeta'_i
\end{split}
\end{equation}
being $s_i$ all possible transmitted symbols in a given amplitude range, and $\xi_i, \zeta_i, \xi'_i, \zeta'_i$ the slack variables corresponding to the real and imaginary constraints.
\end{property}

\begin{property}[Spatial Reference primal coefficients] 
A SVM procedure applied to this constrained optimization problem has to minimize~\eqref{eq:power_min}, and it gives 
\begin{equation}
{{\boldsymbol w}}=\sum_i {\boldsymbol R}^{-1} {\varphi(b_i{\bf a}_0)}{\psi_i} \label{eq:weights_SVM_MVDM}
\end{equation}
\end{property}

\begin{property}(Spatial Reference Kernel). 
The application of~\eqref{eq:weights_SVM_MVDM} in~\eqref{eq:power_min} implicitly gives the kernels 
\begin{equation}
K(b_i {\bf a}_0,b_j {\bf a}_0)={\varphi(b_i{\bf a}_0)}^T{\boldsymbol R}^{-1} {\varphi(b_j{\bf a}_0)},
\end{equation}
which cannot be directly used because an expression for ${\boldsymbol R}$ is not available in infinite dimension RKHS. A kernel eigenanalysis introduced in~\cite{Scholkopf_98b} leads to 
\begin{equation}
K(b_i {\bf a}_0,b_j {\bf a}_0)={N}{\varphi(b_i{\bf a}_0)}^T{\boldsymbol \varPhi} {\boldsymbol
K}^{-1}{\boldsymbol K}_0 {\boldsymbol \varPhi}^T{\varphi(b_j{\bf a}_0)},
\label{eq:kernel_SPProc}
\end{equation}
where ${\boldsymbol \varPhi}$ is a matrix containing all the incoming data used to compute the autocorrelation matrix ${\boldsymbol R}$, and ${\boldsymbol K}_0$ is a kernel matrix containing {all dot products $\varphi(s_n^0{\bf a_0})^\top\varphi(s_m^0{\bf a_0})$.} These kernels can be used to solve a dual problem equal to the one of Property~\ref{Prop:Complex}.
The primal coefficients can be expressed as 
\begin{equation}
{\boldsymbol w}=N{\boldsymbol \varPhi}{\boldsymbol K}^{-1}{\boldsymbol K}_0 {\boldsymbol \psi}
\end{equation}
where $\psi_n = \eta_n + j \nu_n$ are complex-valued dual coefficients. 
\end{property}

\subsection{RSM Application Examples}

This section illustrates two RSM applications, namely, ARX system identification, and spatial reference for antenna array processing. 

\paragraph{SVM  Nonlinear System Identification}

The performance of SVM  with RSM for non-linear system identification was benchmarked in~\cite{Martinez06a}. We used different kernel combinations, namely, separated kernels for input and output processes (SVM-ARX$_{2K}$), accounting for the input-output cross-information (SVM-ARX$_{4K}$), and different combinations of non-linear SVR and SVM-ARX models, all of them with the RBF kernel.

\begin{figure}[t]
\centerline{{\IG[width=8cm]{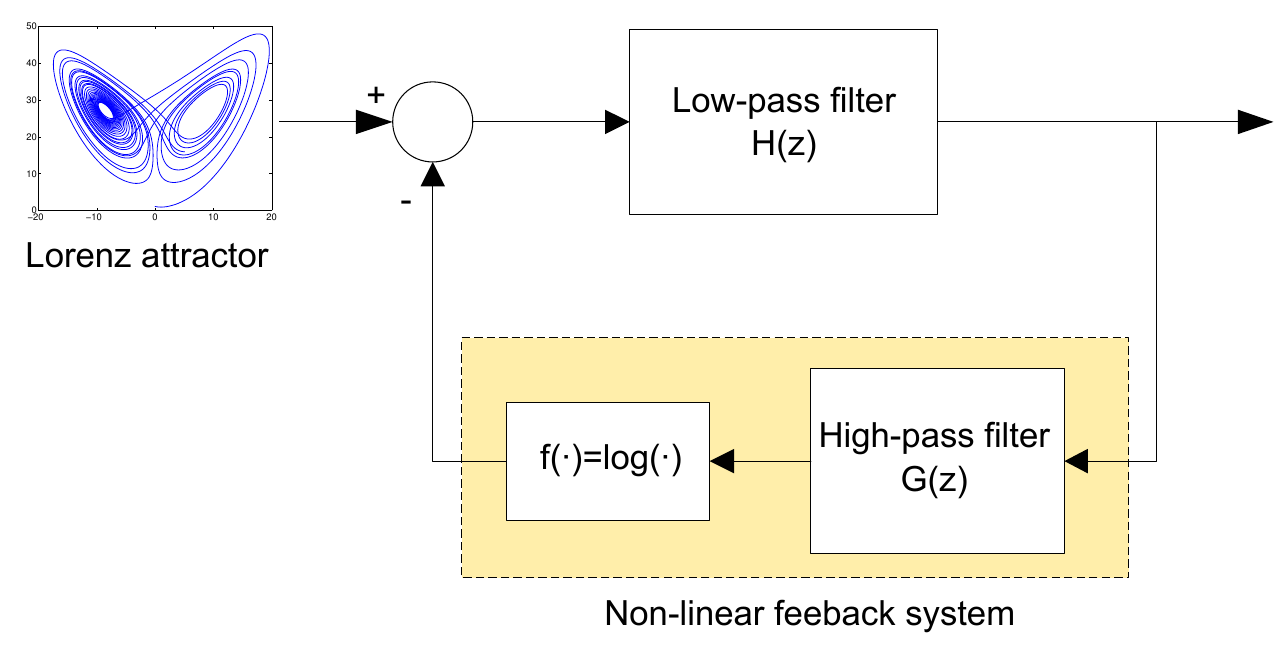}}}
\vspace{-0.3cm}
\caption{Application example for Nonlinear System Identification. System that generates the input-output signals to be
modeled in the SVM  non-linear system identification example.}
 \label{esquema}
\end{figure}

In the global system generating the data (see Fig.~\ref{esquema}), the input discrete-time
signal to the system was generated by sampling the Lorenz system, given by differential equations $
dx/dt = -\rho x + \rho y$, $dy/dt = -xz +rx -y$, and $dx/dt = xy-bz$,
with $\rho=10$, $r=28$, and $b=8/3$ for yielding a chaotic time series. Only the $x$ component
was used as input signal to the system, and it was
then passed through an 8th-order low-pass filter $H(z)$ with
cutoff frequency $\omega_n=0.5$ and normalized gain of -6dB at
$\omega_n$. The output signal was then passed through a feedback
loop consisting of a high-pass minimum-phase channel, given by
$o_n = g_n - 2.01 o_{n-1} - 1.46 o_{n-2} - 0.39 o_{n-3}$,
where $o_n$ and $g_n$ denote the input and the output signals to the channel. Output $o_n$ was distorted with $f(\cdot)=log(\cdot)$.

{We generated 1000 input-output sets of observations and these were split into a cross-validation dataset
(free parameter selection, $100$ samples) and a test set (model performance, following
$500$ samples). The experiment was repeated $100$ times with randomly
selected starting points, and the free parameters were adjusted with cross-validation
in all the experiments. Table~\ref{tab:results} shows the averaged results. 
The best models were obtained when combining SVR and
SVM-ARX models, though no numerical differences were
observed between SVR+SVM-ARX$_{2K}$ and SVR+SVM-ARX$_{2K}$.
In this example, all models considering cross-terms in the
kernels significantly improved SVR results.

\begin{table}[t]
\begin{center}
\small \caption{Mean error (ME), mean-squared error (MSE), mean
absolute error (MAE), and correlation coefficient ($r$) of models in
the test set.}
\begin{tabular}{|l|c|c|c|c|}
\hline
  & {\bf ME} & {\bf MSE} & {\bf MAE} & {$\boldsymbol{r}$} \\
\hline
 {\bf SVR} 						& 0.05& 30.37 & 4.63 & 0.76\\ \hline 
 {\bf SVM-ARX$_{2K}$} & -0.21 & 39.77 & 5.11 & 0.94 \\ \hline 
 {\bf SVM-ARX$_{4K}$} 	& 2.95 & 20.64 & 2.99 & 0.96 \\ \hline 
 {\bf SVR + SVM-ARX$_{2K}$}  & -0.00 & 0.01  & 0.07 & 0.99 \\ \hline
{\bf SVR + SVM-ARX$_{4K}$} & 0.03 & 0.02  & 0.11 & 0.99 \\
\hline
\end{tabular}
\label{tab:results}
\end{center}
\end{table}

\paragraph{Spatial and Temporal Antenna Array Kernel Processing}

The kernel temporal reference (SVM-TR) and spatial reference (SVM-SR) array processors have been benchmarked with their kernel LS counterparts (kernel-TR and kernel-SR), with the linear with temporal reference (MMSE), and with spatial reference (MVDM)~\cite{Martinez06a} . A Gaussian kernel was used in all processors. The scenario consisted of a multiuser environment with one desired three interfering users. The modulated signals were independent QPSK, and the noise was assumed to be thermal, simulated by additive white Gaussian noise. The desired signal was structured in bursts containing 100 training symbols, followed by 1000 test symbols. Free parameters were chosen in the first experiment and fixed.

\begin{figure}[t!]
\begin{center}
\small
\begin{tabular}{cc}
{\bf SVM} & {\bf LS} \\
\includegraphics[width=4cm]{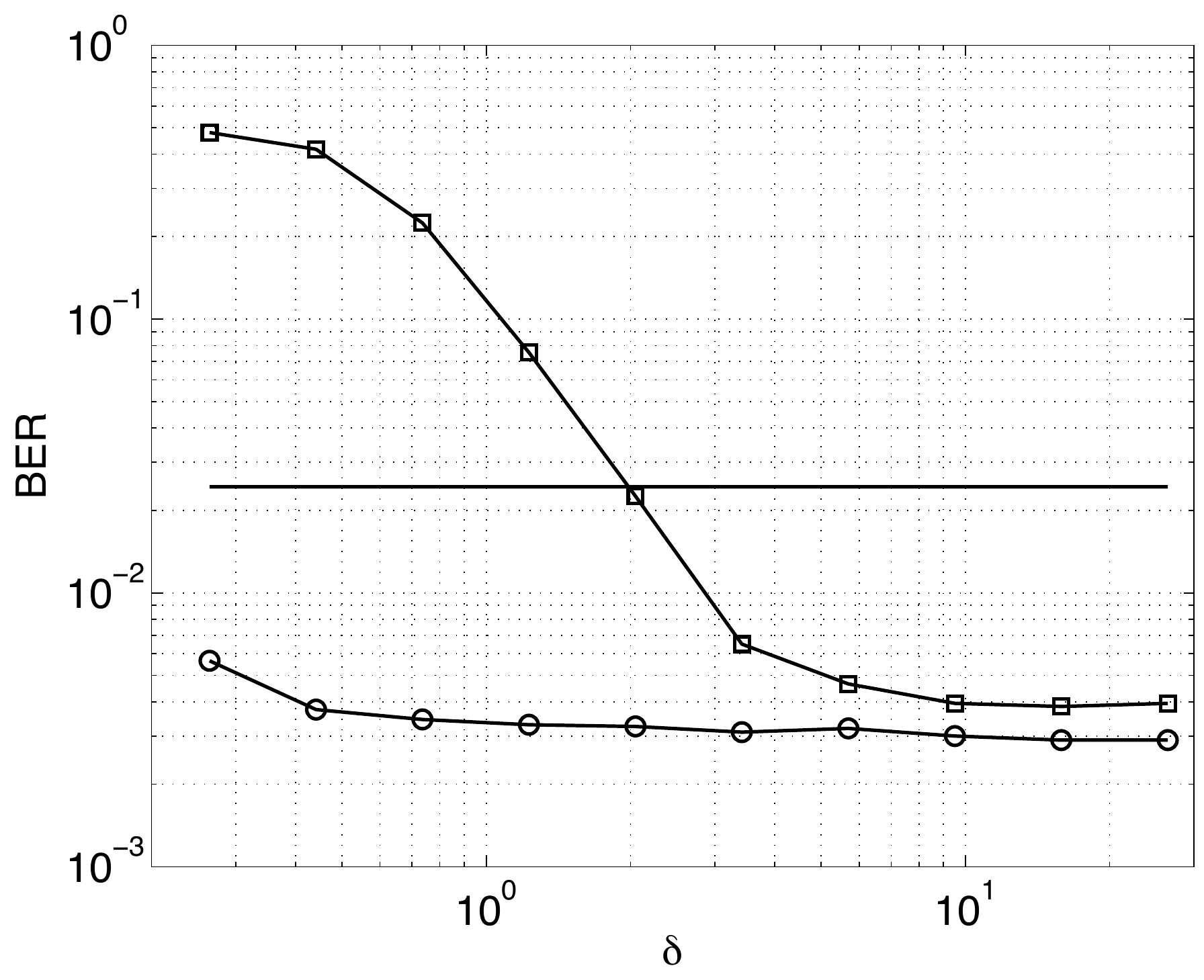} 
 & \includegraphics[width=4cm]{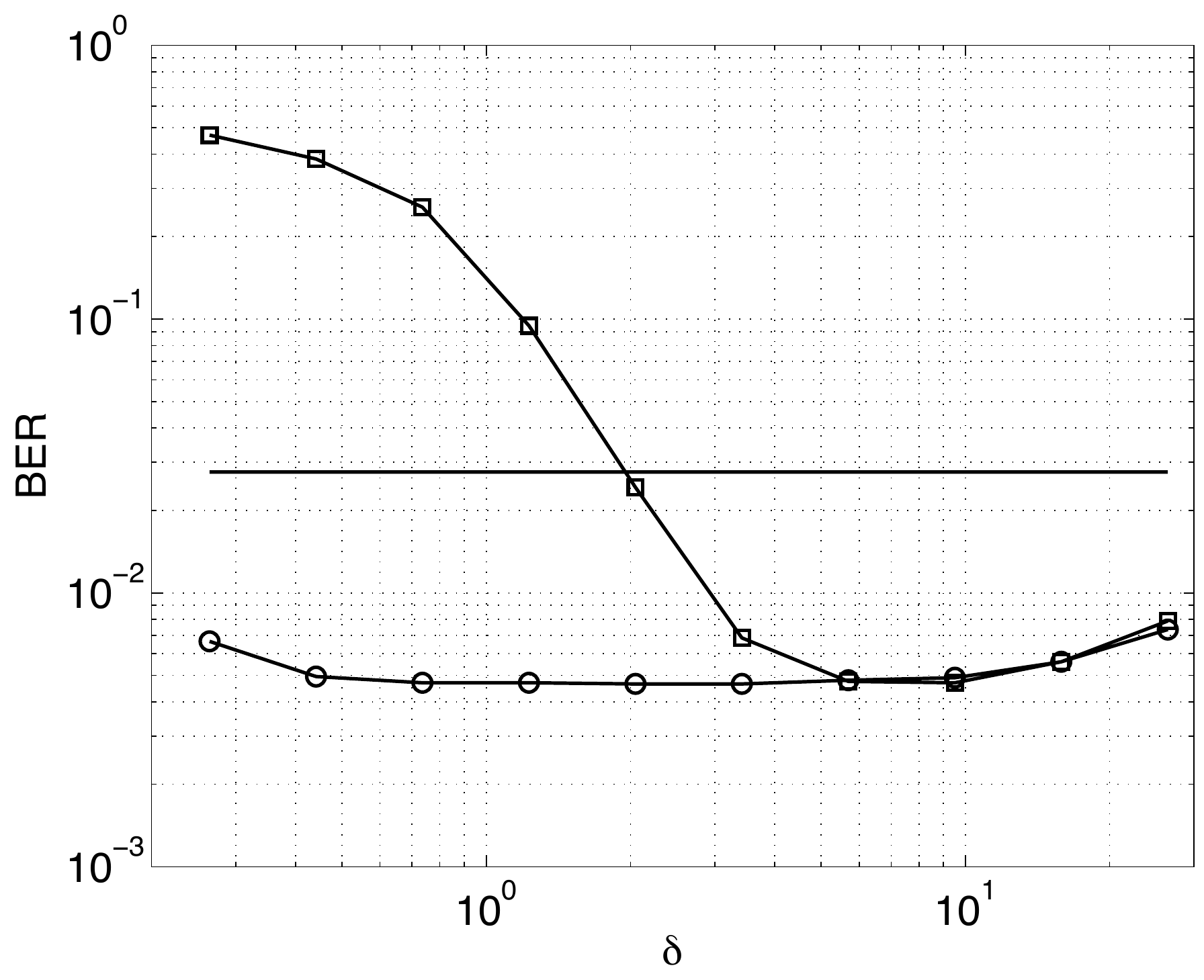} \\
\includegraphics[width=4cm]{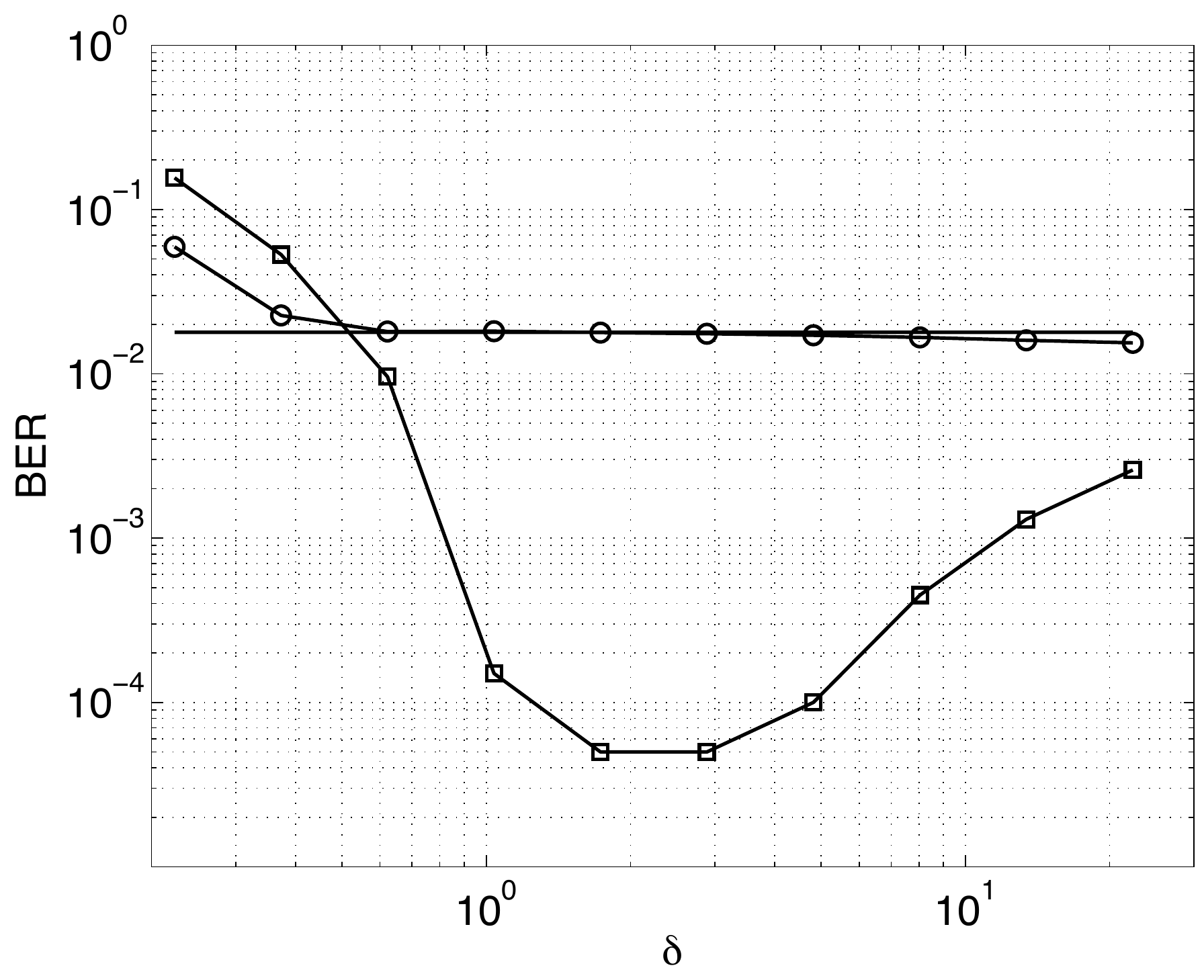} 
 & \includegraphics[width=4cm]{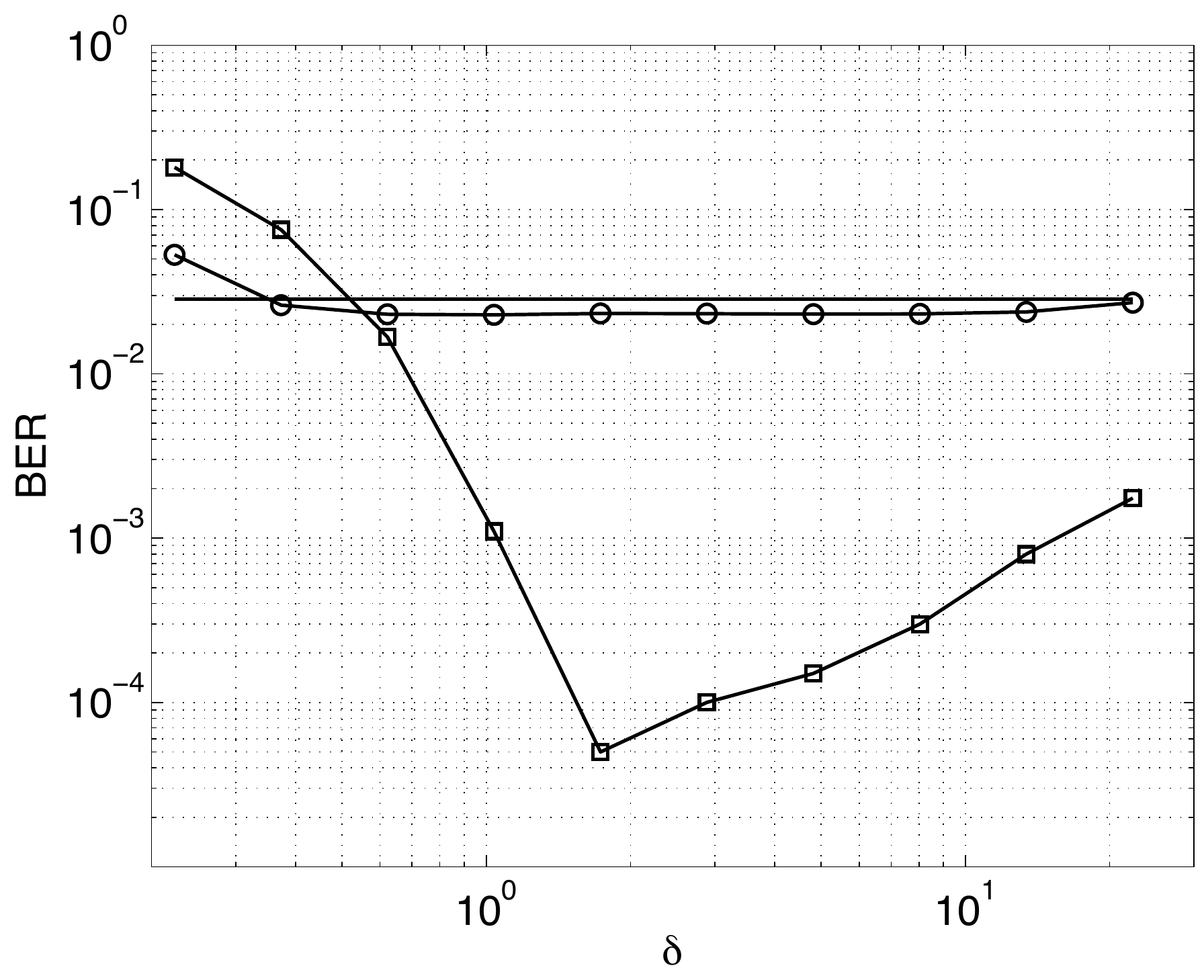}
\end{tabular}
\end{center}
\vspace{-0.5cm}
\caption{Application example for Spatial and Temporal Antenna Array kernel processing. BER performance, as a function of Gaussian RBF kernel
parameter $\delta$, of the TR (squares) and the SR (circles)
in an array of 7 (top) and 5 (bottom) elements and with three interfering signals.
Continuous line corresponds to the performance of the linear
algorithms.}\label{antenas}
\end{figure}

\begin{figure}[t!]
\centering
\includegraphics[width=7cm]{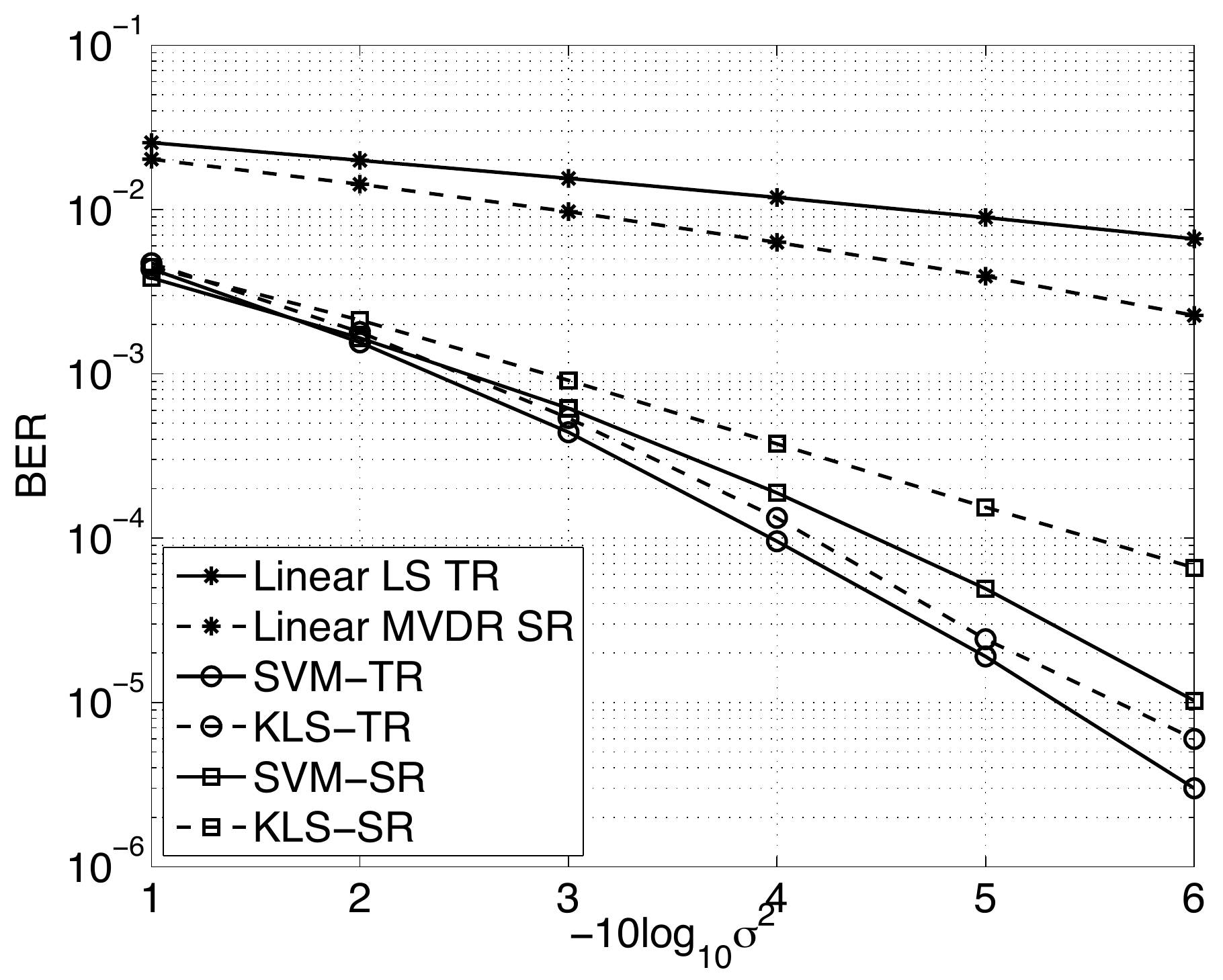}
\vspace{-0.5cm}
\caption{BER performance as a function of thermal noise power for
linear algorithms, SVM SVM-TR, SVM-SR, KLS-TR and KLS-SR.}
\label{fig:BERvsSNR}
\end{figure}

In the first experiment, the BER is measured as a function of kernel parameter $\delta$ for arrays of $5$ and $7$ elements, in an environment of three interferences from angles of arrival of $10^o$, $20^o$ and $-10^o$, and unitary amplitudes, while the desired signal came from an angle of arrival of $0^o$ with the same amplitude as the interferences. Results in Figure~\ref{antenas} show the BER as a function of the RBF kernel width for the temporal and spatial reference SVM algorithms (i.e. SVM-TR and SVM-SR). These results are compared to the temporal and spatial kernel LS algorithms (i.e., KLS-TR and KLS-SR), and for $7$ and $5$ array-elements. The noise power is of $-1$ dB for $7$ elements and -6 dB for $5$ elements. 

The results of the second experiment are shown in Fig.~\ref{fig:BERvsSNR}. The experiment measured the BER of the four non-linear processors as a function of the thermal noise power in an environment with three interfering signals from angles of arrival of $-10^o$, $10^o$ and $20^o$. Desired signal direction of arrival was $0^o$. Performances are compared to the linear MVDR and MMSE algorithms. In this experiment, temporal reference algorithms show a performance slightly better than spatial reference ones. All non-linear approaches show an improvement of several decibels with respect to the linear algorithms. In particular, SVM approaches show better performance than non-linear LS algorithms, with lower test computational burden due to their sparseness properties.

\section{Type III Algorithms: Dual Signal Models}		\label{dual}

An additional class of non-linear {SVM algorithms for DSP} can be obtained by considering the non-linear regression of the time lags or the time instants of the observed signals and using an appropriate choice of the Mercer's kernel. This class is known as {\it DSM based SVM algorithms}. Here, we summarize this approach and pay attention to the interesting and simple interpretation of these SVM algorithms under study in connection with Linear System Theory (see~\cite{Rojo07a, Rojo08a} for details).
{From Table \ref{tab:scheme1}, this is a very natural representation of a signal model stated in SVR form, since the concepts of Mercer's kernel and impulse response can be put together, as will be explained next. In the case of antenna array processing (Table \ref{tab:scheme2}), no DSM algorithm has been proposed yet.}

\subsection{Fundamentals of DSM}

We use the SVR problem statement as support for the algorithm, by making a non-linear mapping of each {\it time instant} to an RKHS, however, the signal model equation of the DSP to be implemented will be the resulting kernel-based solution, by using autocorrelation kernels suitable with the problem at hand. We summarize these ideas in the following theorem. 

\begin{theorem}[DSM Problem Statement] \label{th:dsm}Let $\{y_n\}$ be a discrete time series in a Hilbert space, which is to be approximated in terms of the SVR model in Definition~\ref{prop:SVRmodel}, and let the explanatory signals be just the (possibly nonuniformly sampled) time instants $t_n$ that are mapped to an RKHS. Then, the signal model is given by
\begin{equation}
y_n = y(t)_{|t = t_n} = \langle {\boldsymbol w}, {\boldsymbol \phi}(t_n) \rangle
\end{equation}
and the expansion solution has the following form,
\begin{equation} \label{eq:dsmtheorem}
 {\hat y}_{|t=t_m}(t) = \sum_{n=0}^N \eta_n K^h(t_n,t_m) = \sum_{n=0}^N \eta_n R^h(t_n-t_m)
\end{equation}
where $K^h$ is an autocorrelation kernel originated by a given signal $h(t)$. Model coefficients $\eta_n$ can be obtained from the optimization of~\eqref{eq:primalRaw} (non-linear SVR Signal Model hypothesis in Property~\ref{prop:SVRmodel}), with kernel matrix given by
\begin{equation}
 {\boldsymbol K}^h(n,m) =\langle {\boldsymbol \phi}(t_n), {\boldsymbol \phi}(t_m) \rangle= R^h(t_n-t_m) 
\end{equation}
Hence, the problem is equivalent to non-linearly transforming time instants $t_n$, $t_m$, and making the dot product in the RKHS. For discrete-time DSP models, it is straightforward to use discrete time $n$ for $n^{th}$ sampled time instant $t_n = n T_s$, where $T_s$ is the sampling period in seconds.
\end{theorem}

{Indeed, Eq. \eqref{eq:dsmtheorem} is formally similar to conventional output of SVR  for regression. The conceptually key different is not in the mathematical support given by the non-linear transformation of time instants to the RKHS, but rather that the use of the estimated autocorrelation of the output in terms of the input data stands for a straightforward inclusion of the a priori information of the problem into the support vector formulation. In fact, the use of usual kernels as the RBF can be seen as an easy, yet rough approximation to the autocorrelation of the output as a function of the input variables. 
The autocorrelation kernel concept is allowing to address a number of unidimensional and multidimensional extensions of communications problems from a different and still well founded approach linking DSP and Statistical Learning. In~\cite{Figuera2012},  simple estimations of multidimensional autocorrelation function and their use in a DSM allowed statistically significant advantage in the estimation problem of indoor location from Received Signal Strength in Wifi networks. {In~\cite{Figuera2013}, the connection between DSM with SVM and basic Information Theory concepts like the Wiener filter and the matched filter gave rise to significant improvement in application examples including  bandpass signals or biomedical time series from heart rate variability.}
These are examples that represent partial contributions to the more general problem of building non-linear DSP algorithms with the SVM methodology.}

Theorem~\ref{th:dsm} is used below to obtain the non-linear equations for several DSP problems. In particular, the statement of the sinc interpolation SVM algorithm can be addressed from a DSM~\cite{Rojo07a}, and its interpretation in terms of Linear System Theory allows to propose a DSM algorithm for sparse deconvolution, even in the case that the impulse response is not an autocorrelation.

\subsection{Nonuniform Signal Interpolation with SVM}

The sinc function in the sinc interpolator has a non-negative Fourier transform, and hence it can be used as a Mercer's kernel~\cite{Zhang04} in SVM algorithms.
\begin{property}[DSM for Sinc Interpolation]
Given the sinc interpolation signal model equation in Property~\ref{prop:sincmodel}, and given that the sinc function is an autocorrelation signal, a DSM SVM algorithm can be obtained by using an expansion solution as follows,
\begin{equation}\label{eq:dsmsinc}
 {\hat y}_m = \sum_{n=0}^N \eta_n K(t_n,t_m) = \sum_{n=0}^n \eta_n sinc \left( \sigma_0 (t_n-t_m)\right)
\end{equation}
and Lagrange multipliers $\eta_n$ can be obtained accordingly.
\end{property}
This equation can be compared to the sinc interpolation signal model equation in Property~\ref{prop:sincmodel} in terms of model coefficients and explanatory signals. We can observe that, for uniform sampling, Eq.~\eqref{eq:dsmsinc} can be interpreted as the reconstruction of the observations given by a linear filter, where the impulse response of the filter is the sinc function, and the input signal is given by the sequence of the Lagrange multipliers corresponding to each time instant. A Dirac's delta train can be used for giving a more rigorous proof in the continuous-time equivalent signal model equation. For equally spaced samples, it is easier to see that if we assume that $\{\eta_n\}$ are observations from a discrete-time process, and that $\{K^h(t_n)\}$ is the continuous-time version of an autocorrelation signal (in this case, the {\em sinc} kernel) given by $K^h_n$, then solution $x_n$ can be written down as a convolutional model,
\begin{equation}\label{eq:convolutionaldual}
 x_n = \eta_n\ast K_n.
\end{equation}

\begin{figure}[t!]
\centering
\setlength{\tabcolsep}{2mm}
\begin{tabular}{cc}
\IG[width=4cm]{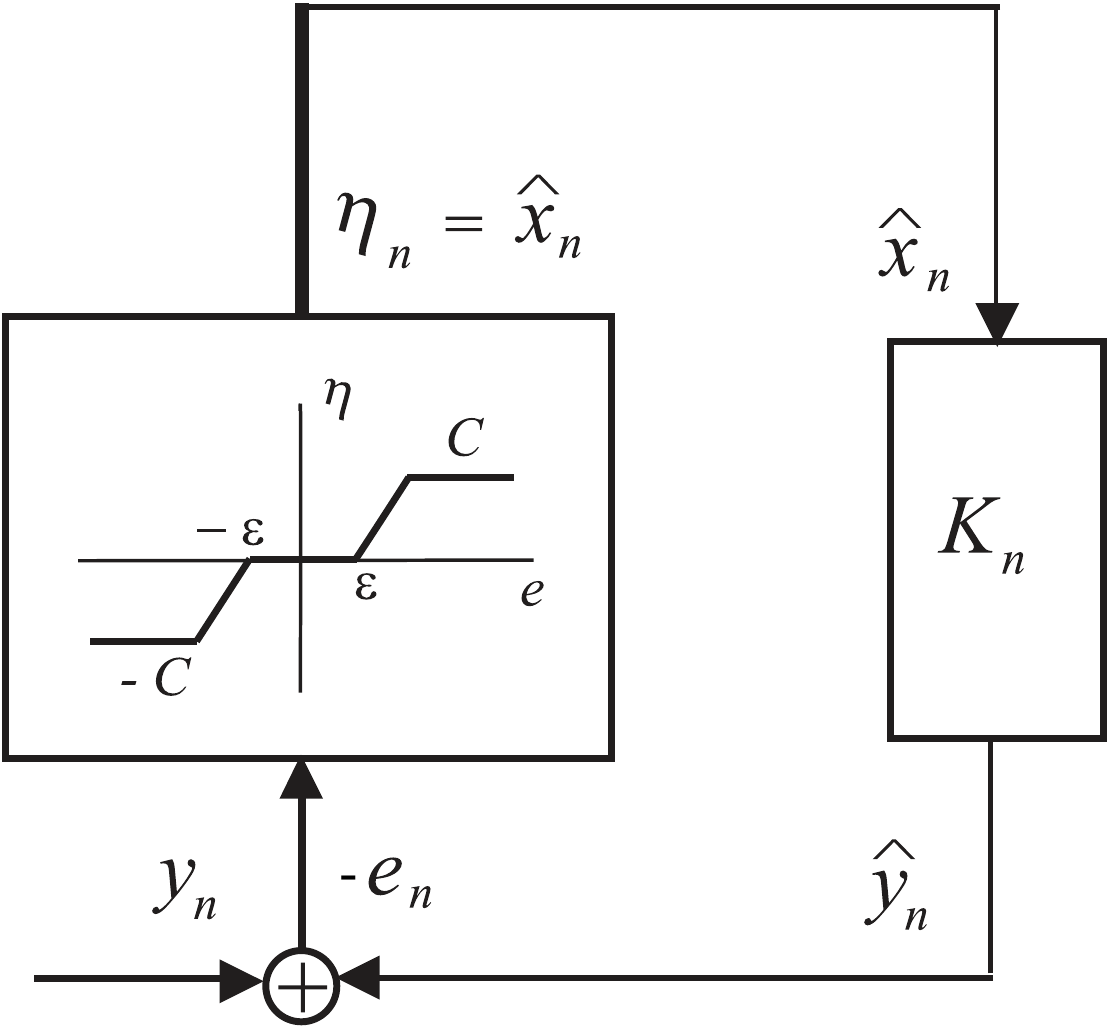} &
\IG[width=4cm]{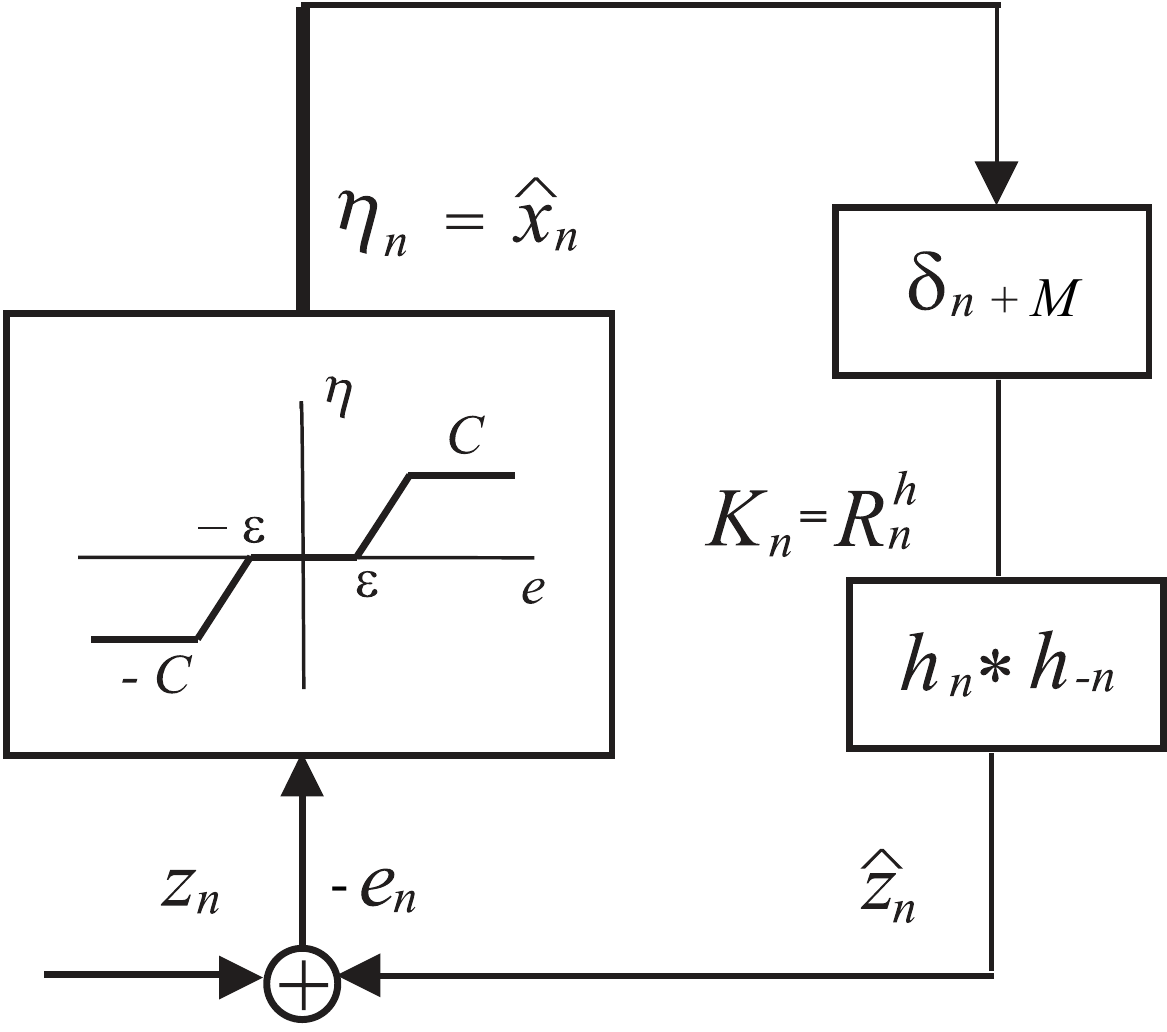} \\
{\fns\it(a)} & {\fns\it(b)} \\
\end{tabular}
 \caption{Signal Model for SVM  sparse deconvolution. (a) Convolutional model for sinc interpolation using the DSM problem statement, where the impulse response of the convolutional model is a valid Mercer's kernel. (b) Convolutional model for sparse deconvolution, for an arbitrary impulse response, where the Mercer's kernel is the autocorrelation of the impulse response signal.}\label{fig:decon_bc}
\end{figure}

Figure~\ref{fig:decon_bc}(a) depicts this situation. This expression is valid as far as we have a valid kernel,
because in this case the impulse response has a nonnegative Fourier transform. However, a non-causal impulse response $h_n$ is used here, which for some applications is not an acceptable assumption. Also, from the point of view of signal processing, the impulse response is constrained to be a valid Mercer's kernel, and as such, it is symmetric, so that this scheme cannot be proposed for every convolutional problem. Nevertheless, by allowing $\varepsilon$ to be non-zero, only a subset of the Lagrange multipliers will be non-zero, thus providing a sparse solution, a highly desirable property in a number of deconvolution problems.

In order to qualitatively compare the {\em sinc} kernel SVM primal and dual signal models for nonuniform interpolation, the following expansion of the solution for the primal signal model approach given in~\eqref{eq:ymodeldiscrete} can be written as follows,
\begin{equation} \label{eq:solprimal}
\begin{split}
\hat y_m = &\sum_{n=0}^N a_n \textrm{sinc}(\sigma_0(t_n-t_m)) \\
& = \sum_{n=0}^N \left(\sum_{r=0}^N\eta_r \textrm{sinc}(\sigma_0(t_m-t_r)
 \right) \textrm{sinc}(\sigma_0(t_n-t_m)).
\end{split}
\end{equation}
Comparison between~\eqref{eq:solprimal} and~\eqref{eq:dsmsinc} reveals that these are quite different
approaches using SVM for solving a similar signal processing problem. For the primal signal model formulation, limiting the value of $C$ will prevent these coefficients from an uncontrolled growing (regularization effect). {For the dual signal model formulation, the SRM principle, which is implicit in the SVM formalism~\cite{Vapnik95}, will lead to a reduced number of non-zero coefficients.}

\subsection{Sparse Signal Deconvolution}

Given the observations of two discrete-time sequences $\{y_n\}$ and $\{h_n\}$, deconvolution consists of finding the discrete-time sequence $\{x_n\}$ fulfilling
\begin{equation}
y_n = x_n \ast h_n + e_n.
\end{equation}
In many practical situations, $x_n$ is a sparse signal, and solving this problem using {an SVM} algorithm can have the additional advantage of its sparsity property in the dual coefficients. If $h_n$ is an autocorrelation signal, then the problem can be stated as the sinc interpolation problem in the preceding subsection, using $h_n$ instead of the sinc signal. This approach requires an impulse response that is a Mercer's kernel, and if an autocorrelation signal is used as kernel (as we did in the preceding subsection for the sinc interpolation), then $h_n$ is necessarily a non-causal linear, time-invariant system. For a causal system, the impulse response cannot be an autocorrelation. A first approach is the statement of the PSM in Section~\ref{primal}. The solution can be expressed as
\begin{equation}
\hat x_n = \sum_{i=0}^N \eta_i h_{i-n}.
\end{equation}
hence, an implicit signal model equation can be written down, which is
\begin{equation}
\hat x_n = \sum^N_{i=M} \eta_i h_{i-n} = \eta_n \ast h_{-n+M} =
 \eta_n \ast h_{-n} \ast \delta_{n+M}.
\end{equation}
This means that the estimated signal is built as the convolution of the Lagrange multipliers with the time-reversed impulse response and with a $M$-lagged time-offset delta function $\delta_n$. Figure~\ref{fig:decon_bc} shows the schemes of both SVM algorithms. According to the Karush-Khun-Tucker conditions, the residuals between the observations and the model output are used to control the Lagrange multipliers. In the DSM based SVM algorithms, the Lagrange multipliers are the input to a linear, time invariant, non-causal system whose impulse response is the Mercer's kernel. Interestingly, in the PSM based SVM algorithms, the Lagrange multipliers can be seen as the input to a single linear, time invariant system, whose global input response is $h^{eq}_n = h_n \ast h_{-n} \ast \delta_{n-M}$ (Fig.~\ref{fig:decon_a}). It is easy to show that $h^{eq}_n$ is the expression for a Mercer's kernel, that emerges naturally from the PSM formulation. This provides with a new direction to explore the properties of the DSM SVM algorithms in connection with classical Linear System Theory, which is following described.

\begin{property}[DSM for Sparse Deconvolution Problem Statement]
Given the sparse deconvolution signal model equation in Property~\ref{prop:sparsedeconmodel}, and given a set of observations $\{y_n\}$, these observations can be transformed into
\begin{equation}
z_n = y_n \ast h_{-n} \ast \delta_{n-M},
\end{equation}
and hence, a DSM SVM algorithm can be obtained by using a expansion solution with the following form,
\begin{equation}\label{eq:soldualsignal}
 {\hat y}_m = \sum_{n=0}^n \eta_n K(n,m) = \eta_n \ast h_n \ast h_{-n} = \eta_n \ast R^h_n,
\end{equation}
where $R^h_n$ is the autocorrelation of $h_n$, and Lagrange multipliers $\eta_n$ can be readily obtained according to the DSM Theorem.
\end{property}
Figure~\ref{fig:decon_bc}(b) depicts this new situation, which can be easily solved now. This simple transformation of the observations allows to address the sparse deconvolution problem for any impulse response $h_n$ to be considered in the practical application.

\subsection{DSM Application Examples}

This section highlights the differences between primal and dual SVM signal model equations, focusing on nonuniform interpolation and sparse deconvolution.

\paragraph{Nonuniform Interpolation}

\begin{figure}[t]
\begin{center}
\setlength{\tabcolsep}{1pt}
\begin{tabular}{c}
\IG[width=6cm]{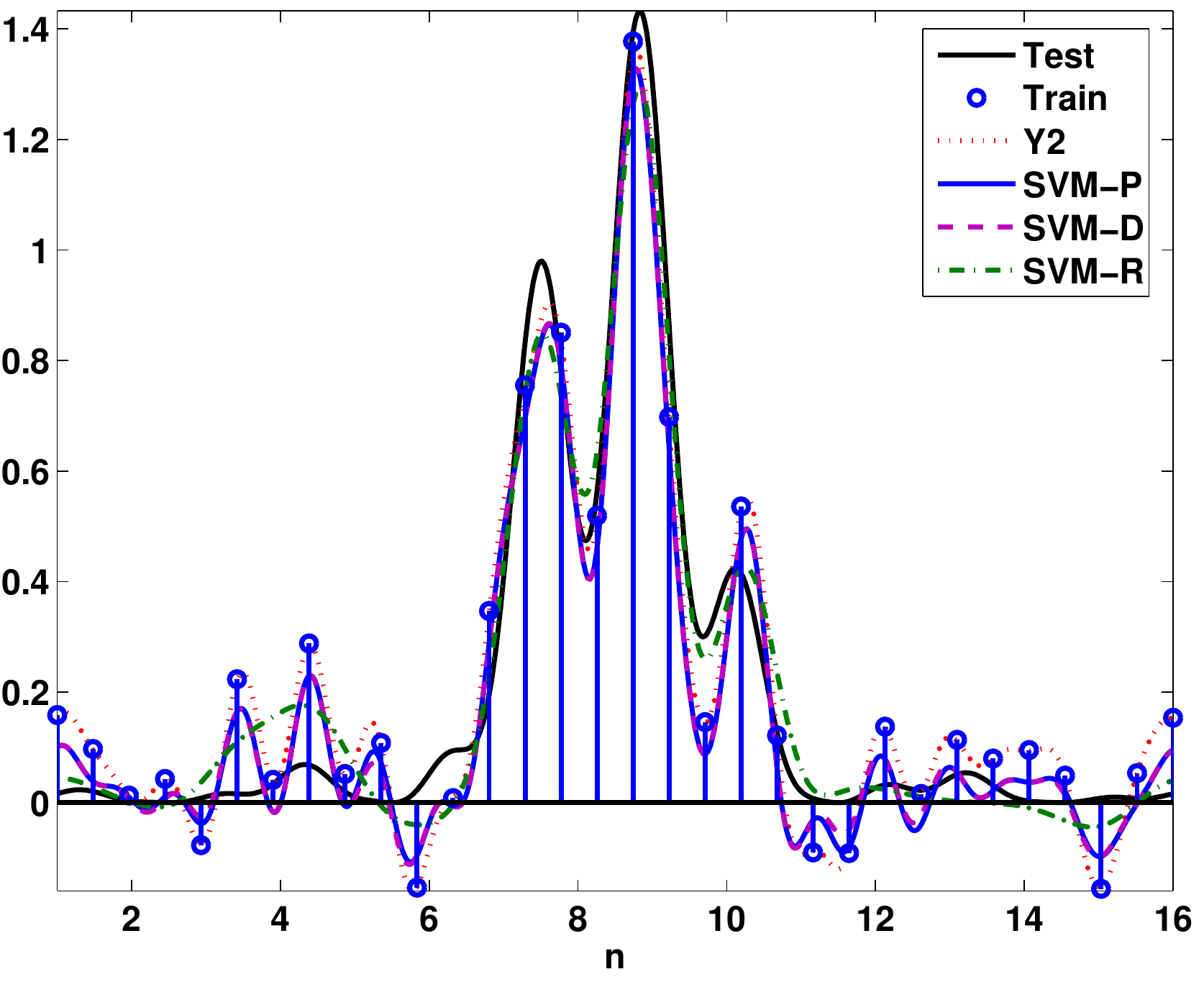} \\ {\fns\it(a)} \\
\IG[width=6cm]{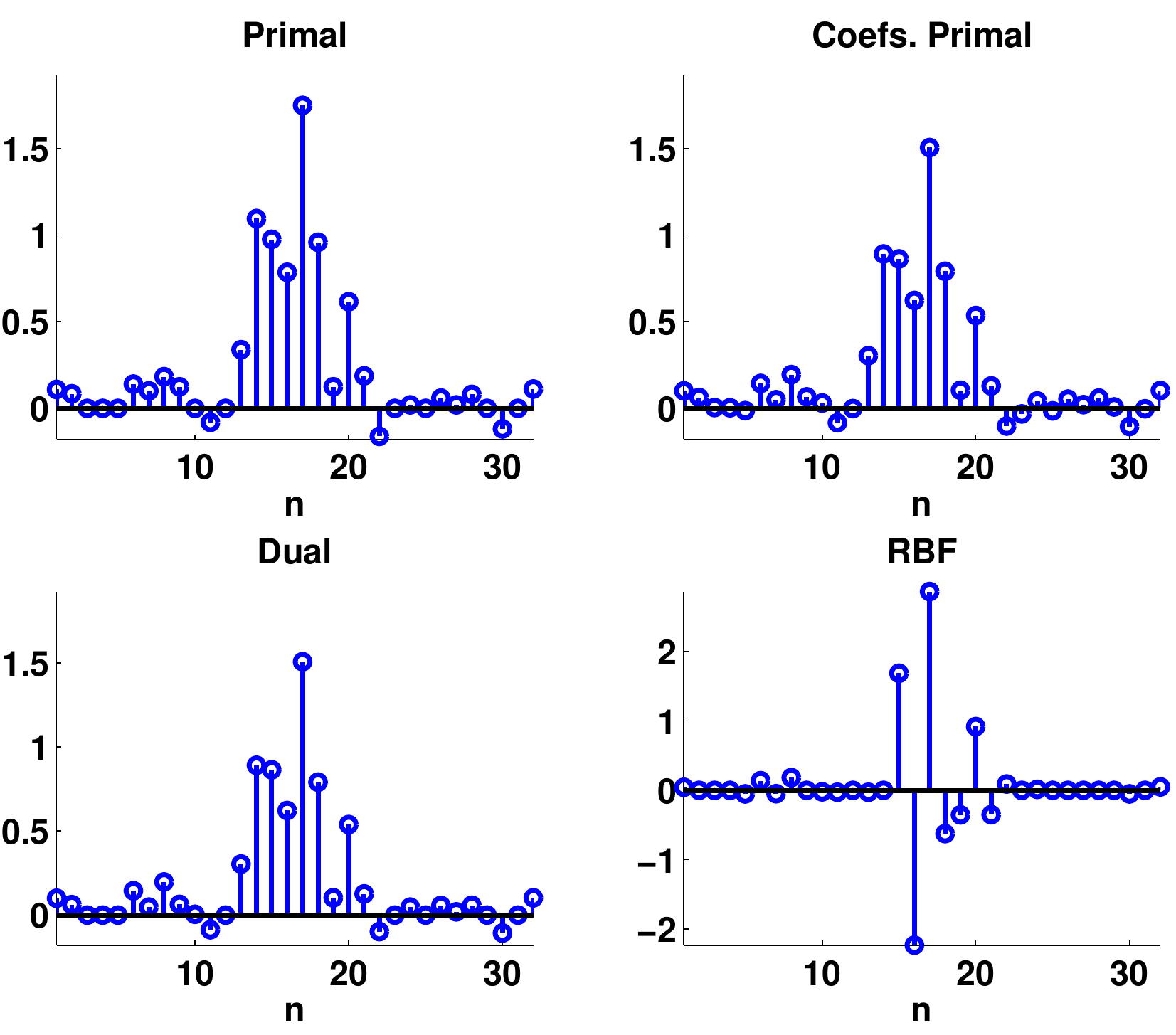} \\ {\fns\it(b)} \\
\end{tabular}
\end{center}
\vspace{-0.5cm}
\caption{Application example for SVM sinc interpolation from PSM and DSM. (a) Training, test, and reconstructed signals in the
time domain. (b) Temporal representation of the model coefficient discrete-time sequences.}
\label{fig:sincExample}
\end{figure}

For comparison purposes, PSM and DSM were used for denoising a signal consisting on the sum of two squared sincs, one of them being a lower level, amplitude modulated version of the baseband component, 
\begin{equation}\label{eq:doublesinc}
y(t) = \textrm{sinc}^2\left(\frac{\pi}{T_0} t \right) \left( 1 + \frac{1}{2} \sin\left( 2\pi f t \right) \right)+ e(t),
\end{equation}
where $f = 0.4$ Hz. A set of $L=32$ samples was used with averaged sampling interval $T=0.5s$~\cite{Rojo07a}. The signal to noise ratio was 10dB, and performance was measured on a test, noise-free, uniformly sampled version of the signal with sampling interval $T/16$, as an approximation to the continuous time signal. Figure~\ref{fig:sincExample} shows the training and test signals, as well as the PSM and DSM interpolation with {\em sinc} kernel, and DSM with RBF kernel. The time representation of the model coefficients was similar in PSM and DSM, but not equal, when using the {\em sinc} kernel. {Note that the sparseness} obtained by the RBF kernel was significantly higher than for the {\em sinc} kernel (see the original reference for details).

\paragraph{Sparse Deconvolution}

In~\cite{Rojo08a}, sparse deconvolution with SVM is benchmarked for analyzing a B-scan given by an ultrasonic transducer array from a layered composite material, details on this application can be found elsewhere~\cite{Olofsson04}. Figure
\ref{fig:true_example}(a) shows a signal example (A-scan) of the sparse signal estimated by several methods. The same panel
also shows the reconstructed observed signal. The $L_1$ deconvolution yielded good quality solution with a noticeably number of spurious
peaks, the DSM algorithm yielded a good quality solution with less spurious peaks, and the Gaussian Mixture (GM) algorithm often failed at detecting the low amplitude peaks. Figure~\ref{fig:true_example}(b) shows the reconstruction of the complete
B-scan data.

\begin{figure*}[t] \footnotesize
 \begin{center}
\setlength{\tabcolsep}{1pt}
 \begin{tabular}{cc}
  \begin{tabular}{cc}
  \IG[width=4cm]{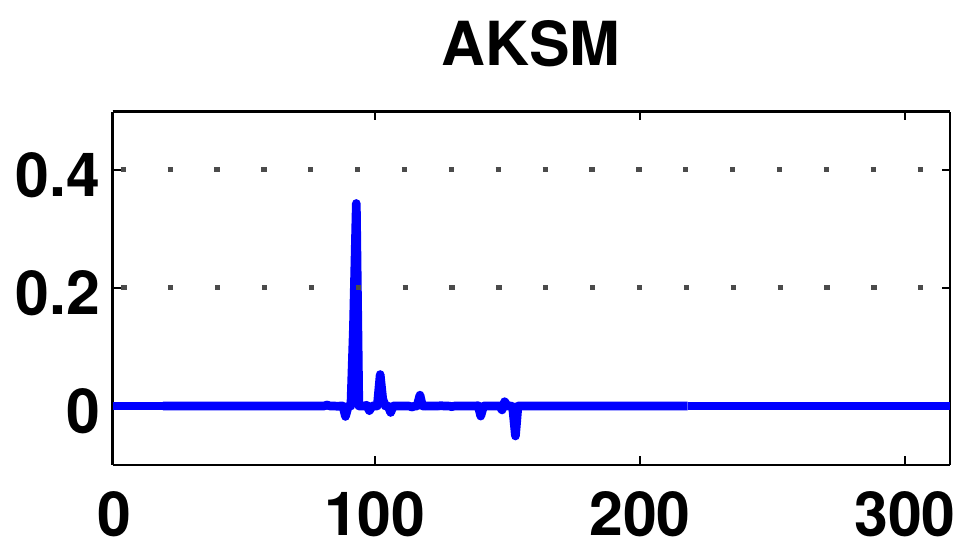} & \IG[width=4cm]{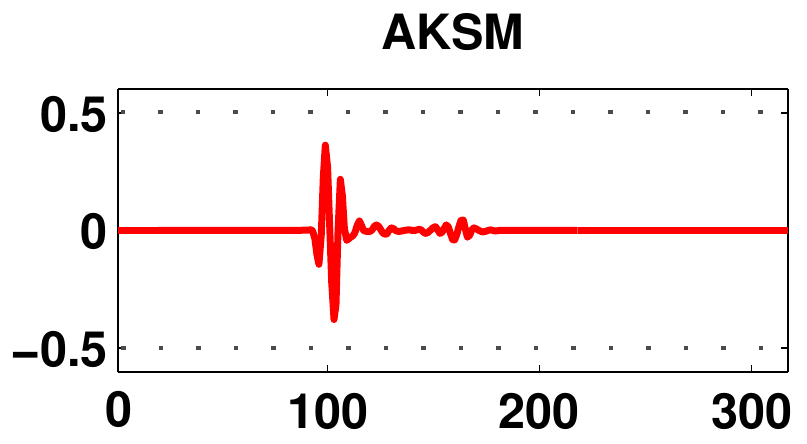} \\
  \IG[width=4cm]{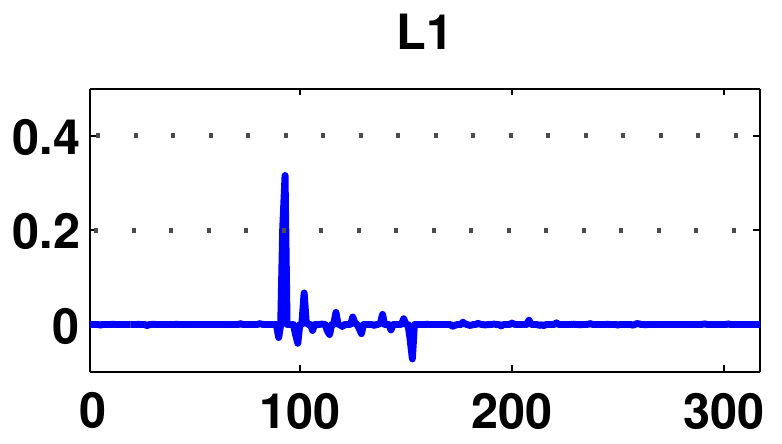}   & \IG[width=4cm]{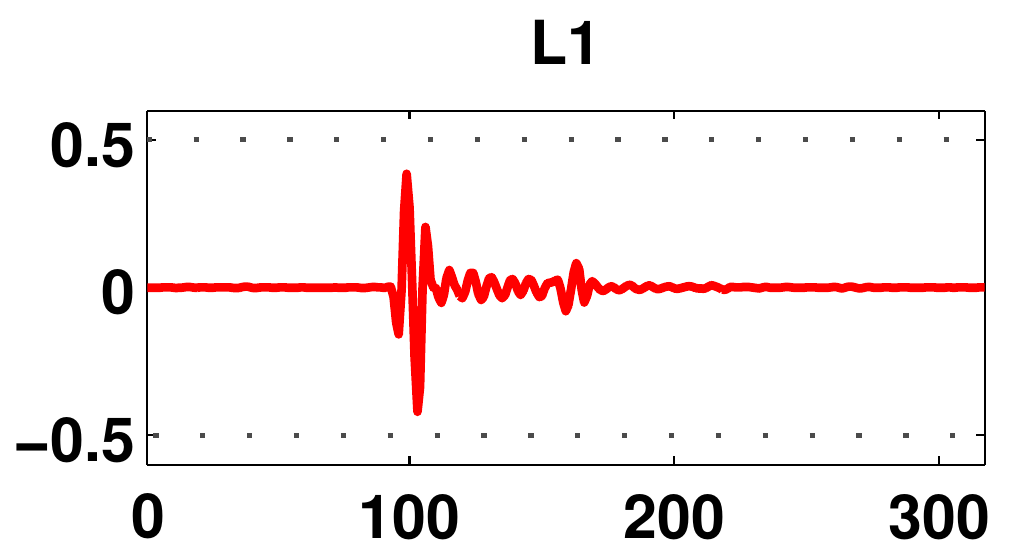} \\
  \IG[width=4cm]{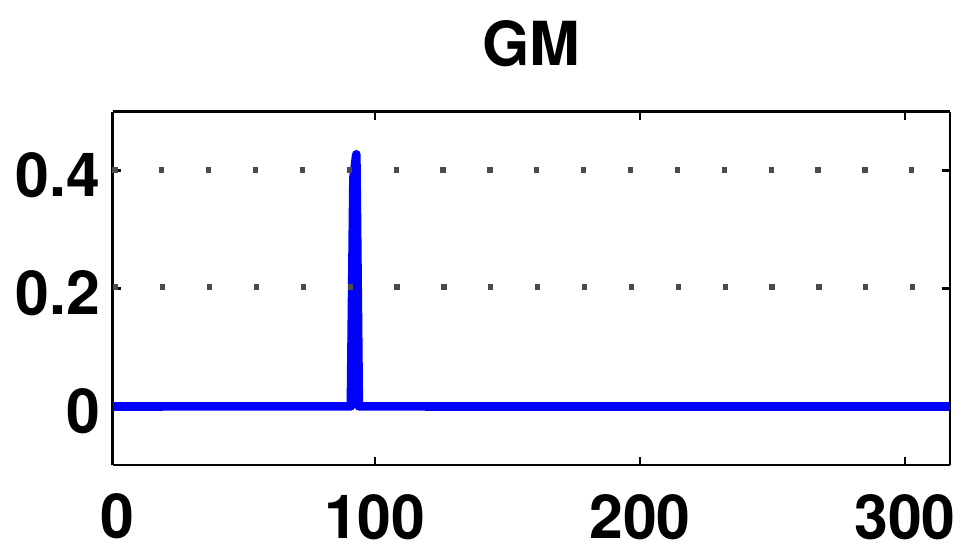}   & \IG[width=4cm]{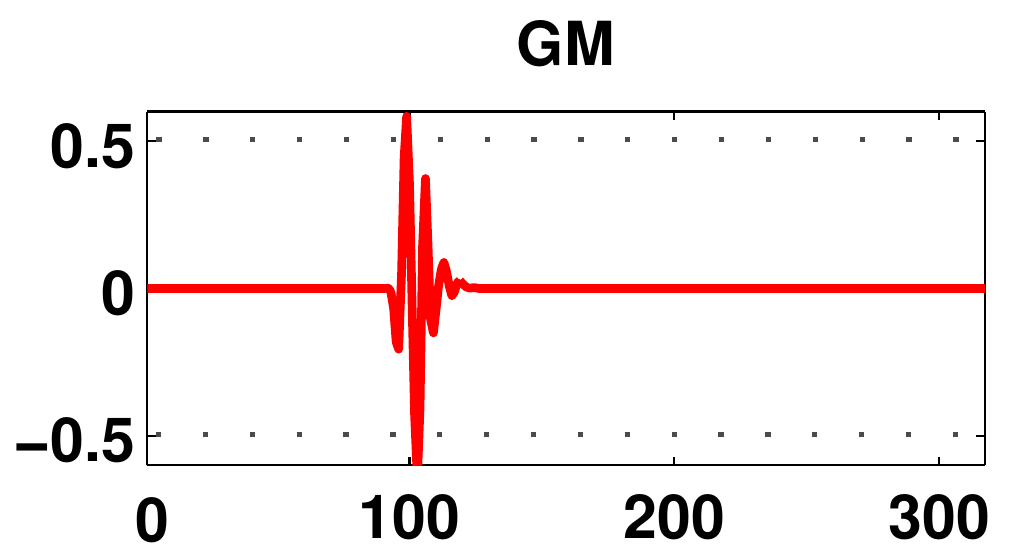} \\
  \end{tabular}
  & 
  \begin{tabular}{cc}
  \IG[width=4cm]{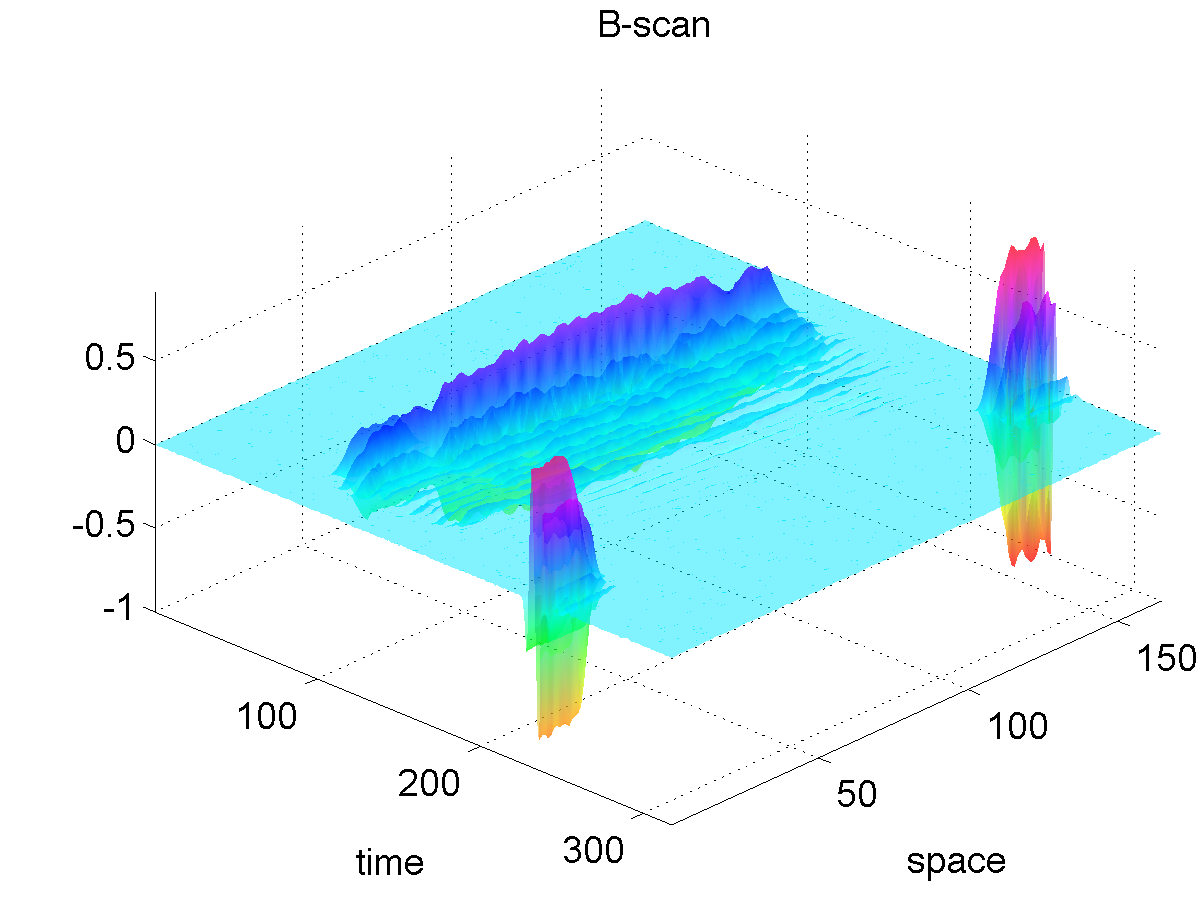} & \IG[width=4cm]{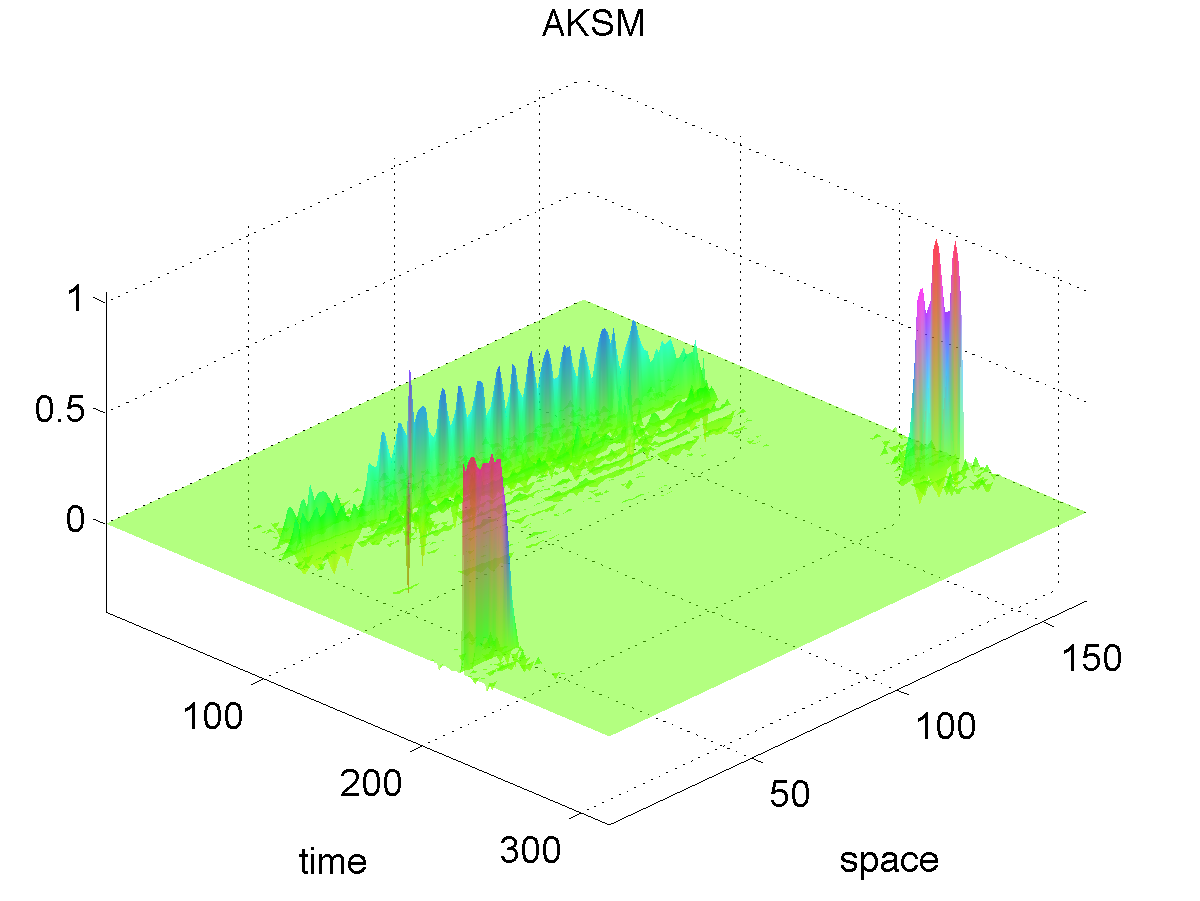} \\
  \IG[width=4cm]{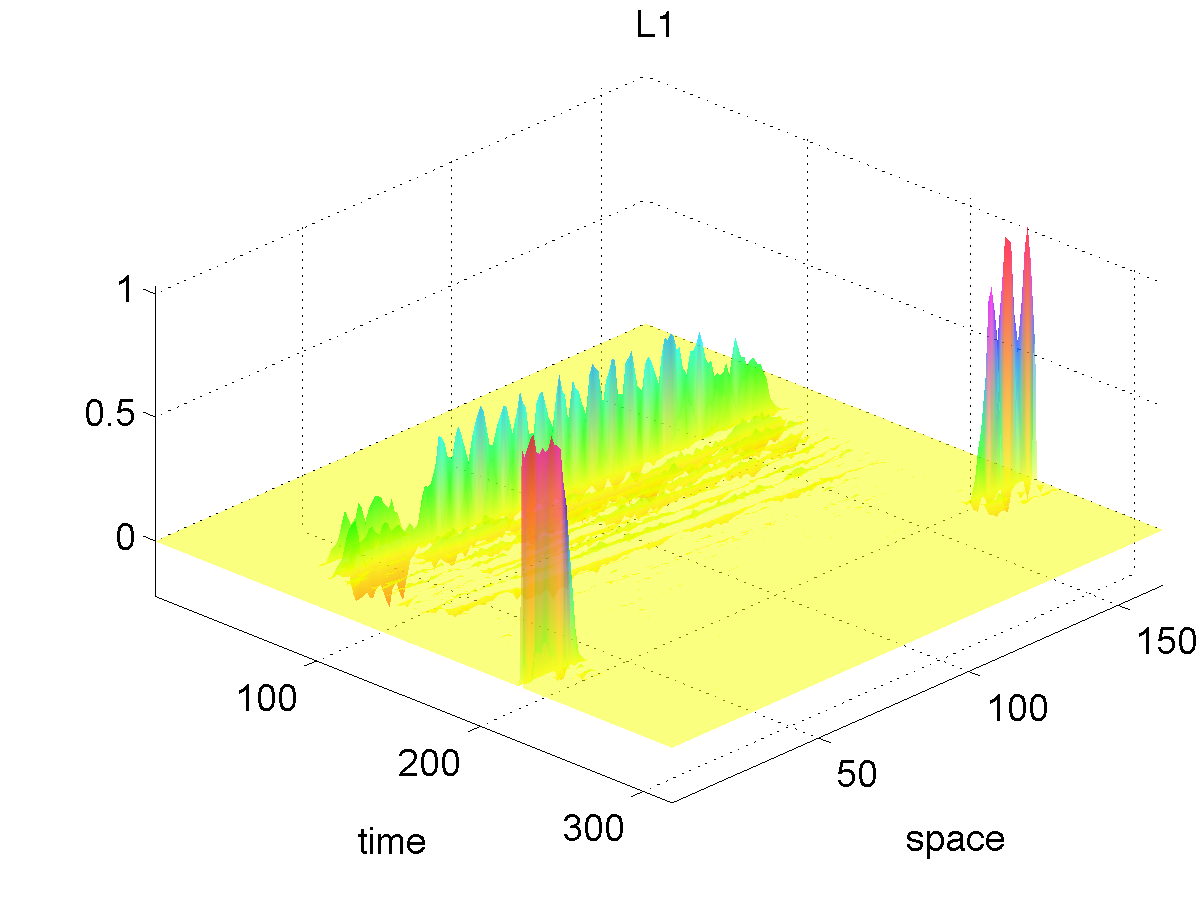}     & \IG[width=4cm]{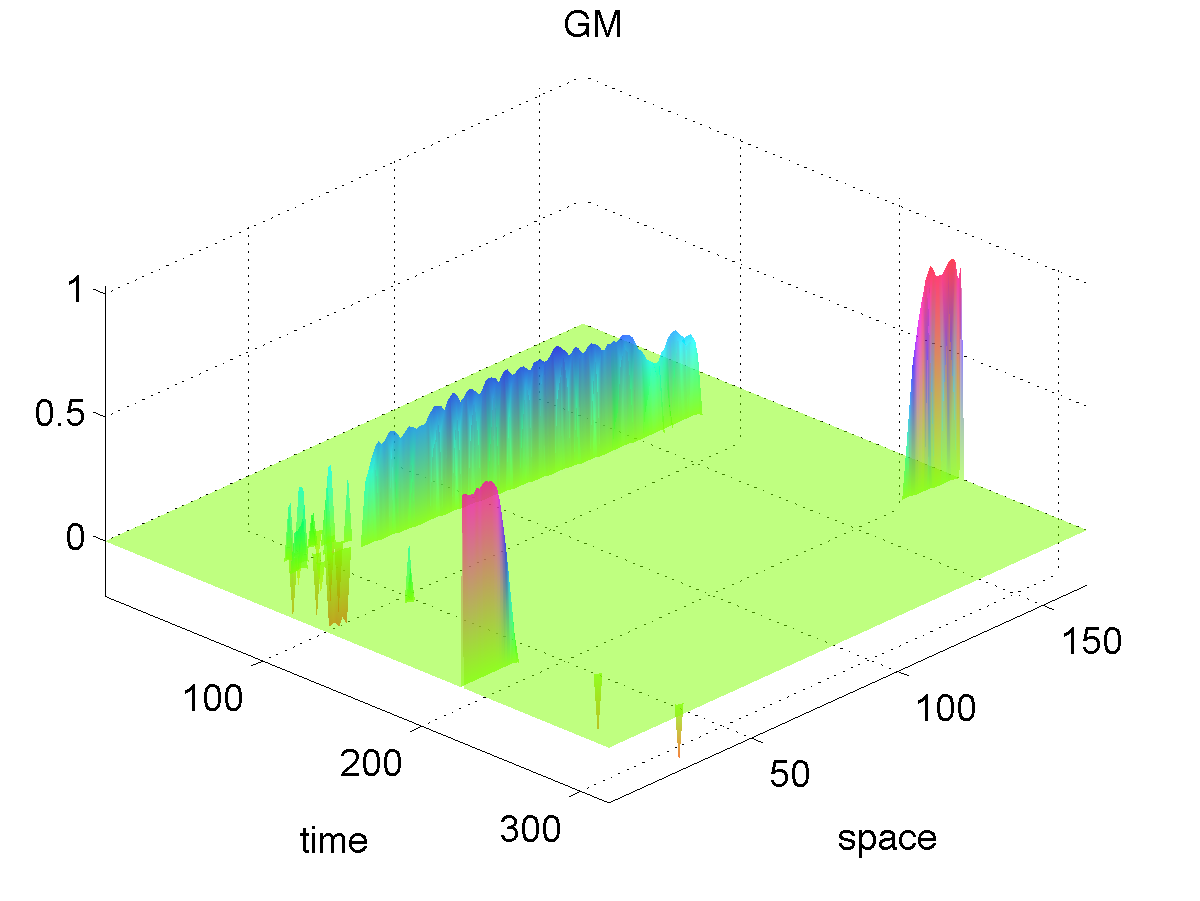} \\
  \end{tabular} \\
  {\fns\it(a)} & {\fns\it(b)} \\
  \end{tabular}
\end{center}
\vspace{-0.5cm}
\caption{Example of real data application: sparse deconvolution of the impulse response in a ultrasound transducer for the analysis of a layered composite material. (a) Example of deconvolution of a single A-scan line with each algorithm: left, sparse
estimated signal; right, estimated A-line. (b) Deconvolution of the B-scan data.}
\label{fig:true_example}
\end{figure*}

\section{Discussion and Conclusions}		\label{conclusions}

{We presented a unified framework for signal processing with SVM. The framework is constituted by a  set of basic {\em tools} (regularization, cost function, primal-dual signal formulations), {\em operations} (shift-invariant Mercer's kernel, {\em sinc} kernels, direct sum of Hilbert spaces), and {\em signal models} (PSM, RSM, DSM).} For illustration purposes, several standard problems in DSP have been addressed, namely filtering, linear and non-linear  system identification, spectral estimation, interpolation, sparse deconvolution, and antenna array processing. The capabilities of this framework for  developing new models for DSP have been  illustrated in selected experiments, with improved  results compared to standard formulations.

The main motivation behind the presented framework relies on the lack of insight of the signal-model structure when using the standard SVR.  Certainly, developing signal processing tools with support vector machines  has led to the extensive use of the kernel trick for reformulating the linear problem at hand. This is, of course, a valid approach to develop SVM based methods for signal processing. However, as highlighted both theoretically and experimentally, this can be a limited methodological approach as a wide variety of time series structures are not taken into account. The statements of a signal model equation in the primal equations, in the RKHS, and in the solution (dual) equation of the SVM, yield different algorithms. Also, by inspecting the signal structure, it is possible to develop more appropriate SVM models that accommodate it, and that generalize the common stacked-vector approach plus ``kernelization''. The framework has demonstrated to be particularly useful to develop SVM formulations for time-series data.

Another relevant aspect for developing SVM-based signal processing algorithms is the use of a flexible cost function. We make emphasis here on the use of a  general $\varepsilon$-Huber cost function which generalizes several SVR losses. In fact,  the Vapnik's $\varepsilon$-insensitivity cost~\cite{Smola04} is a particular  case for $\delta=0$, and the LS-SVM cost~\cite{Suykens00} is a particular case for $\varepsilon=0$ and $\delta C$ large enough. 

The proposed framework for DSP is quite general and admits any kind of non-parametric regression machine,  such as Gaussian Processes (GPs)~\cite{Rasmussen04}, for  the same task. Certainly, the nice properties of  GPs would be worth exploring in the near future. In particular, free parameter  tuning through the maximization of the marginal likelihood would make the  methods more appealing. Also, providing predictive variances could be  an interesting advantage for some signal processing applications, such as system identification or time series prediction. 

Future work includes the possibility of developing schemes for input feature selection in time-series problems, and the extension of these concepts to multi-dimensional  problems, specially for interpolation in three-dimensional spaces.  Implementing efficient time recursion in the RKHS to develop working kernel versions  of standard adaptive and recursive algorithms in DSP is also a field to be explored.

\section*{Acknowledgments}

This paper has been partially supported by the Spanish Ministry for Education and Science under project TEC2010-19263 and the  Spanish Ministry of Economy and Competitiveness (MINECO) under project  TIN2012-38102-C03-01. 

\bibliographystyle{unsrt}
\small
\bibliography{superbibclean}

\end{document}